\documentclass[pdflatex,sn-basic]{sn-jnl} % "pdflatex" only stops the class loading breakurl, which XeTeX cannot use
\usepackage{iftex}
\ifPDFTeX % submission systems often compile with pdfLaTeX; ko.TeX supports it (kotex-utf with Type 1 Korean fonts)
\usepackage[utf8]{inputenc}
\usepackage{textcomp}
\else
\usepackage{fontspec}
\fi
\usepackage{graphicx,booktabs,tabularx,url,amssymb,amsmath}
\usepackage{kotex} % Korean terms in Section 2 and Appendix B
\usepackage{newunicodechar}
\newunicodechar{≥}{\ensuremath{\geq}}
\newunicodechar{−}{\ensuremath{-}}
\newunicodechar{→}{\ensuremath{\rightarrow}}
\newunicodechar{κ}{\ensuremath{\kappa}}
\usepackage[title]{appendix}
\begin{document}
\title[LexTemporis]{Grading Executable Legal Rules Against Independent Records: Transitional Provisions and Korean SME Status}

\author*[1]{\fnm{Jinhyung} \sur{Bae}}\email{kevinbae2006@hufs.ac.kr}
\affil*[1]{\orgdiv{Department of Industrial and Management Engineering}, \orgname{Hankuk University of Foreign Studies}, \orgaddress{\city{Yongin}, \country{Republic of Korea}}}

\abstract{Laws are increasingly turned into programs that decide cases, such as whether a firm was a small or medium enterprise (SME) on a given date, and large language models (LLMs) now write them. The programs are usually graded against answers written by someone who read the same law, so a program can score well by sharing the grader's reading rather than by getting the law right. LexTemporis grades them instead against records of what actually happened: a released test bed of 10,901 firm-years of SME status disclosed by Korean listed companies, with every version of the governing rules. An appendix adds comparison cases from US income tax and German hospital minimum volumes.
\par
Graded against the disclosures, the author's own rule shows where the date changes outcomes: mainly outside the main statutory text. Modelling one transitional provision raises agreement from 83–86\% to 89–90\%. The disclosures also test the author's reading of a contested provision, favouring it in 130 of 189 firm-years where the readings diverge; hand checks and a blind second annotator confirm these labels, and a convention fitted to the disclosures holds on held-out firms.
\par
Four LLMs also wrote rules from different selections of the law. The pre-registered comparison of these retrieval conditions found no condition better across the four models. In exploratory analyses, the advantage of supplying the right provisions is smaller against the records than against the author's reading, partly by construction. Grading legal programs against the grader's own reading can overstate how well they work.}

\keywords{Executable legal rules, Rules as code, Large language models, Evaluation, Transitional provisions, Temporal validity}

\maketitle

\section{Introduction}

Was company X a small or medium enterprise (SME) when it filed its annual report in March 2017? Banks pricing loans, agencies awarding subsidies, and the firm itself must answer this question, and under Korean law the answer requires at least five provisions in four layers: the statute, the enforcement decree, the decree's annexes, and the addenda (Table 1 in Section 2). Article 2 of the addenda to the 2014 amendment keeps firms that were SMEs under the old rules in that category until 31 March 2018. A firm that exceeded the 2015 ceilings but had been an SME under the old rules was still an SME in March 2017; from April 2018 it either ceased to be one or entered a three-year grace period, depending on how Paragraph 2 of that article is read. The disclosures lean towards the reading under which it ceased to be one (Section 6.2), without settling the question. None of this appears in the main text of the statute.

Recent work addresses parts of this problem: retrieving the version of a provision in force on a date [Cymbler 2026; Prior et al. 2026; Akarajaradwong et al. 2026; Li et al. 2026], following references from statutes to subordinate rules [Ju \& Lee 2026; Chae et al. 2026; An et al. 2024], and having LLMs translate statutes into code [Chen \& Tanaka-Ishii 2026; Ahmad et al. 2026; Han et al. 2026; Dou et al. 2026]. Independent answer keys are not new either: Law\&Order and TaxCalcBench grade tax computation against independently produced returns, and the Axiom Foundation compares encoded rules with TAXSIM [Axiom Foundation 2026]. What is missing is an evaluation of status decisions that turn on dates, delegation, and transitional provisions, across many years, against keys that the person who read the law did not write.

In terms of legal theory, the question is intertemporal: which norm's effect governs a decision on a given date (Section 2). Formal models of legal time state which times of a norm a rule must track; they do not measure how often an encoding that tracks the wrong one changes a decision, nor whether its author read a transitional provision correctly. Keys produced without reference to the author's reading can. Our claim is that executable rules for such decisions should be graded against keys independent of the reading that produced them. Such keys can change what one concludes about three things: how far the written rule determines decisions and through which layer of the law the date enters; whether the author's own reading holds; and how different ways of supplying the law to a model compare.

\textbf{Two references, two roles.} Every synthesis result is reported twice. The reference rule is one explicit reading of the law, written by the author; agreement with it measures whether a synthesized rule reproduces that reading. The answer key records what the system's own decision-makers did, without reference to our reading; agreement with it measures whether a rule reproduces practice. The reference rule's own agreement with the key bounds how far the written rule, as read, determines the decisions. A difference between conditions that appears against the reference rule but not against the key is a property of the reading, not of the law's effect on decisions.

\textbf{Why these cases.} We chose the cases by three criteria, not by the importance of their domains: (i) within the observed years the outcome depends on the date, through amendments or transitional provisions; (ii) the decision needs provisions from more than one layer (statute, delegated rules, annexes or delegated documents, addenda); (iii) an answer key not written by the person who read the law exists at scale. Many legal domains meet (i) and (ii); (iii) is rare. Korean SME status meets all three: the size test was overhauled in 2015, with a transitional addendum running to 2018, and its ceilings were raised in 2025; and listed firms state their status in every annual report, which gives 10,901 labelled firm-years. The status carries tax, finance, and procurement benefits, but here it is a test item: we do not aim to build a better SME classifier. The comparison cases meet the same criteria with a different kind of key and a different layer carrying the date: in US income tax the amounts change every year through delegated Revenue Procedures and the key is an established calculator; in German minimum volumes the threshold sits in a transitional clause and the key is the insurers' decisions. These cases test the claim beyond one statute and one kind of key. Because the paper's questions turn on the Korean case, we report them in Appendix F and refer to them where they bear on it.

We ask two questions.

\begin{itemize}
\item \textbf{RQ1 (what the key shows about the written rule).} How far does a rule written from the law explain an independent answer key, through which layer does the date enter, and does the key confirm the author's own reading of the law?
\item \textbf{RQ2 (what the key changes in evaluating synthesis).} Do conclusions about LLM-synthesized rules, in particular which retrieval condition is better, change when the author's reference rule is replaced by the independent key? This is a baseline evaluation, not a test of a hypothesis that our retrieval wins; beyond one pre-registered comparison, its results are exploratory, and RQ1 carries the paper.
\end{itemize}

\textbf{Contributions.}

\begin{enumerate}
\item \textbf{LexTemporis, a released test bed} that grades as-of-date status decisions involving delegation and transitional provisions, over many years, against answer keys independent of the author's reading: at its centre 10,901 Korean firm-years with every version of the act, decree, annexes, and addenda, graded against the firms' own disclosures; comparison test beds in US income tax (2,000 TAXSIM-graded returns) and German hospital minimum volumes (7,845 decisions with 24 rule versions; Appendix F); and scoring scripts into which other researchers can plug their own systems. Earlier validations checked human-built implementations against official calculators or test suites (Section 3). To our knowledge, no earlier test bed grades executable rules, across many law versions, against independent records of real outcomes and an established external calculator; the closest work uses single official outcomes [Coppola 2026] and official test suites for one code base [Merigoux et al. 2021b].
\item \textbf{What the independent key shows about the written rule} (RQ1), on the Korean case:

\begin{itemize}
\item \emph{the times of the norm become measurable}: in Korea the date enters through transitional and status-fixing provisions more than through the annex amounts (Section 6.1); in the comparison cases it enters through yearly delegated amounts (US) and through a transitional clause and an evaluative review standard (Germany), which illustrates, on four years of decisions, where rules as code should stop, a limit so far argued only conceptually [Yermukhametova et al. 2026] (Appendix F.3);
\item \emph{the author's reading is itself tested}: the disclosures favour the author's reading of a contested transitional addendum in 130 of 189 firm-years where the readings diverge, without settling the question, and hand checks and a blind second annotator confirm the labels; the convention for the start of the grace period, fitted to the disclosures for the reference rule, holds in a split-sample check on held-out firms (Section 6.2).
\end{itemize}
\item \textbf{Exploratory observations on LLM synthesis} (RQ2): the pre-registered comparison of retrieval conditions finds no condition better across four models; for three of the models, the lead of our retrieval shrinks when it is measured against the key instead of the reference rule (Section 6.3).
\end{enumerate}

The baseline system that produces the synthesized rules (delegation-aware retrieval and a verify-and-repair loop driven by properties of the legal text) and an addenda reader designed to improve the retriever are released with the test bed as instruments of the evaluation, not claimed as contributions (Section 5; Appendices A and D.5). The test bed (1) and what the key shows (2) carry the paper whatever the LLM baseline shows; we do not claim that LLMs perform better with our retrieval.

The name refers to the principle that an act is governed by the law in force at its time (\emph{lex temporis}; \emph{tempus regit actum}). The benchmark operationalizes it for executable status rules; it is not a doctrinal treatment of intertemporal law.

\begin{figure}[t]
\centering
\includegraphics[width=\linewidth]{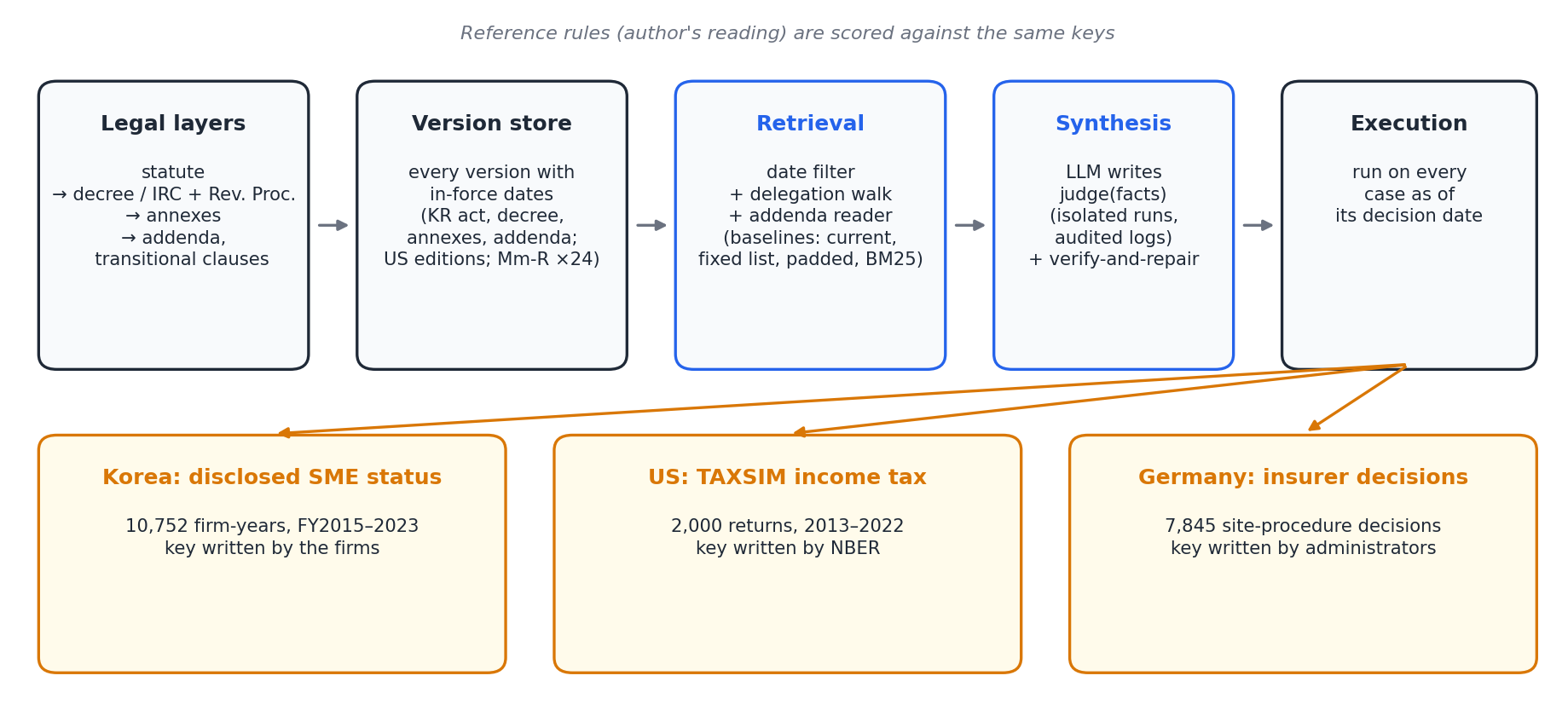}
\caption{The test bed. Each case is decided as of its own date from a store of every version of the governing texts; rules written by an LLM, and the author's reference rules, are scored against keys that the author did not write.}
\end{figure}

\section{Background: the Korean SME rule}

\begin{table}[t]
\caption{Layers of the Korean SME rule.}
\footnotesize
\begin{tabularx}{\linewidth}{p{0.22\linewidth}XXX}
\toprule
Layer & Instrument & What it decides & Changes in the study window \\
\midrule
Statute & Framework Act on SMEs, Art. 2 & Definition; grace period (§2(3)) & – \\
Decree & Arts. 3, 7, 7-2 & Size tests: 3-year average revenue (Art. 7), total assets < KRW 500 bn (Art. 7-2), independence & 2015 overhaul from employee/capital to revenue tests \\
Decree & Art. 3-3 & Status applies for one year from three months after the fiscal year end & – \\
Decree & Art. 9 & Grace period granted once only & – \\
Annex & Annex 1 & Revenue ceiling per industry (KSIC code) & 2015 new table; 2017 KSIC reclassification; 2025 ceilings raised for 16 of 44 industry rows \\
Annex & Annex 2 & Aggregation of affiliated firms' revenue & – \\
Addenda & 2014.4.14 Art. 2 (addenda of the amendment that made the 2015 overhaul: promulgated 2014, in force 1 January 2015) & Firms that were SMEs under the old rules stay SMEs until 2018-03-31 and are not deemed SMEs under the grace period of Act Art. 2(3); whether this excludes grace after 2018-03-31 is contested (Section 6.2) & expires 2018-03-31 \\
Addenda & 2025.9.1 Art. 2 & Special rule for the 2025 raise & from 2025-09-01 \\
\bottomrule
\end{tabularx}
\end{table}

\textbf{Addenda and their temporal scope.} Every Korean amending act or decree ends with addenda, of the same rank as the instrument, that state its temporal scope: an effective-date article, application rules (적용례) that say to which facts or periods the new rule applies, transitional measures (경과조치) that preserve the old legal position for a period, and special rules (특례). The 2014 addendum is a transitional measure. It does not amend the definition; it keeps an old status alive for a fixed time, and its effect ends on a date written into it. Addenda are not merged into the amended text: the consolidated text published by the National Law Information Center keeps every past addendum at its foot, so the current decree still carries the 2014 addendum long after its effect ended. Whether a reader, or a retriever, treats an addendum as live on a given date is a question of reading its temporal scope, not of finding it. The Center also publishes every historical version with its effective date and machine-readable delegation metadata.

\textbf{Official interpretations.} The Ministry of Government Legislation has issued dated interpretations on the provisions this study depends on: the grace period (15-0157, 2015; 21-0131, 2021), the application of the 2014 amendment (14-0128, April 2014), affiliated firms and independence (06-0089, 09-0003, 11-0251, 11-0343, 13-0371, 20-0344), and the definition (12-0579, 15-0465). The administration generally follows them, but they do not bind the courts (Appendix B.2).

\textbf{Times of the norm.} Legal theory distinguishes a norm's being in force (its membership in the legal system) from its applicability: whether courts and officials must apply it to a case. Applicability is relative to the time of the facts, so a norm no longer in force can remain applicable to facts that occurred while it was in force (ultractivity), and a norm can be in force before it becomes applicable [Bulygin 2015; Navarro \& Moreso 1997]. A transitional deeming achieves the same effect for an encoding by a different device: a norm in force attaches the old status to later facts by a legal fiction. Formal models of legal change represent these times explicitly: when a norm is enacted, is in force, and produces effects, and the time of the facts to which it applies, including retroactive change [Palmirani \& Brighi 2006; Governatori \& Rotolo 2010; Tamargo et al. 2019; Moguillansky et al. 2023]. Intertemporal law starts from \emph{tempus regit actum} and from the distinction between a new law's immediate effect on situations still in progress and its retroactive effect on completed ones [Roubier 1960]. Transitional provisions depart from that starting rule on purpose: they keep an earlier legal position for a stated period, apply a new norm only to later facts, or fix a status for a period.

\textbf{Transitional provisions in Korean law.} The drafting standards of the Ministry of Government Legislation treat the addenda as the instrument's provisions for the transition [Ministry of Government Legislation 2017]. Two constitutional doctrines frame them (Appendix B.2 gives the holdings). First, protection of legitimate expectations: the Constitutional Court weighs reliance on the old rule against the purpose of the change [Constitutional Court 1995], and expectations beyond the boundary the legislature has drawn for a transitional provision are not protected reliance [Constitutional Court 2019]; the 2014 addendum is the legislature's own answer to that weighing. Second, an addendum's own temporal life can be contested: the Supreme Court kept an old addendum to a tax act alive after a whole amendment, a reading the Constitutional Court held unconstitutional [Supreme Court 2008; Constitutional Court 2012]. Whether an addendum is still applicable on a date is therefore a legal question, not a lookup. It is also the question that an executable rule for an as-of-date decision must decide. For the 2014 addendum the question is concrete: after 31 March 2018 a covered firm still above the ceilings either receives the three-year grace period (reading (a)) or does not (reading (b)); Section 6.2 tests the two readings against the disclosures. Table 2 classifies the temporal structures the three cases contain and what the independent keys measure for each.

\begin{table}[t]
\caption{Temporal structures in the three cases: what an encoding that tracks only the text in force gets wrong, and what the keys measure.}
\footnotesize
\begin{tabularx}{\linewidth}{p{0.22\linewidth}XXXX}
\toprule
Type & Temporal structure & Instance & What an in-force-only encoding gets wrong & Measured (against the key unless stated) \\
\midrule
Transitional deeming & an old position stays applicable to facts after the new norm is in force, until a stated date & Korea: 2014 decree addenda, Art. 2 (in force 1 January 2015; deeming until 31 March 2018) & judges covered firms by the new ceilings from 2015 & modelling it with the once-only rule raises agreement by 5.1 points in FY2015–2017 (Section 6.1) \\
Lagged status & status for a period is fixed by facts of the preceding period & Korea: Decree Art. 3-3 & applies a new annex to the period in which it enters into force & none of 44 firms between the old and the 2025 ceilings switched in the FY2025 report (outside the key) \\
Anticipatory delegated parameter & an instrument is in force before the period it governs & US: Revenue Procedure for tax year Y, issued in the autumn of Y−1 & takes the latest document before the decision date, which governs the next year & the previous year's law changes tax by 1\% or more on 50\% of returns (Appendix F.3) \\
Embedded schedule & the current text states year-specific values & Germany: Mm-R transitional clauses & nothing, if the current text is read whole & functions from the current and the date-correct versions score identically (Appendix F.4) \\
Evaluative standard & applicability turns on a judgement under an indeterminate concept & Germany: refutation on "justified substantial doubts" & the judgement cannot be encoded & the share of entitled below-threshold sites rose from 0.34 to 0.64 (Appendix F.3) \\
\bottomrule
\end{tabularx}
\end{table}

Appendix B defines the terms used in this paper for readers from law and from computer science; Appendix B.2 gives the legal background in detail for readers outside Korea.

\section{Related work}

\textbf{Rules as code.} Hand-built executable law runs from the British Nationality Act in Prolog [Sergot et al. 1986] to Catala, OpenFisca, and PolicyEngine-US, which maintains multi-year US tax rules by hand against TAXSIM [Feenberg \& Coutts 1993]; time-aware norm representations exist [Governatori \& Rotolo 2010; Palmirani et al. 2011]. These systems are written by experts; we grade rules derived from retrieved text against keys outside the system.

\textbf{LLM statute-to-code.} LLMs translate statutes into Prolog [Holzenberger et al. 2020; Jurayj et al. 2026; Dou et al. 2026], Catala [Lorenzo et al. 2025; Ahmad et al. 2026], or Python, and codify retrieved statutes and precedents into executable explanation traces for judgment predictions [Chen \& Tanaka-Ishii 2026]. Law\&Order [Han et al. 2026] repairs US tax programs against OpenTaxSolver returns and evaluates them on independently authored TaxCalcBench returns, for one tax year, without statute versions, delegated documents, or transitional rules, and with reference outputs rather than properties of the law as the repair signal. Synedrion [Gogani-Khiabani et al. 2026] builds tax software for tax year 2021 against a hand-written reference, with LLM-inferred metamorphic relations driving repair. It neither varies the tax year nor selects the delegated Revenue Procedure, and it does not ask whether a relation encodes one reading of the law. Such metamorphic testing of LLM statutory reasoning was proposed by Padhye (2024). Ahmad et al. detect changes in a German reference graph but do not evaluate against versions in force on a date. Building-code checking does the same translation for BIM models [Zheng et al. 2023; Madireddy et al. 2025]. In this line of work, grading uses researcher-written labels or reference implementations.

\textbf{Temporal and delegation-aware retrieval.} In FiscalQA Pro, static retrieval-augmented generation retrieves the correct version of French tax provisions 0\% of the time [Cymbler 2026]; German [Prior et al. 2026], Chinese [Li et al. 2026; Fan et al. 2026], Thai [Akarajaradwong et al. 2026], and Korean [Jung et al. 2026] benchmarks test identification of the temporally valid norm; TALD-Bench finds nine LLMs strongly biased towards the newer law on 51,956 Chinese civil judgments around the Civil Code [Huang et al. 2026], and LawShift benchmarks legal judgment prediction under statute shifts [Han et al. 2025]; and SAT-Graph RAG studies versioned retrieval over the Brazilian constitution qualitatively [de Martim 2025]. Delegation and reference links serve retrieval in Korean regulations [Ju \& Lee 2026; Chae et al. 2026] and criminal law, over a mention graph of 192,974 nodes [An et al. 2024], and are extracted in Japanese law [Fujita et al. 2026]. Point-in-time statutes are produced by executing amendment language [Mador-Haim \& Hershowitz 2023] or by LLMs, which consolidate German statutes exactly in only half of single steps [Prior et al. 2025]. Transitional clauses in EU legislation have been classified with hybrid AI [Vagnoni \& Palmirani 2025]; we parse Korean addenda article by article for as-of-date retrieval. Time in legal reasoning goes back at least to Vila \& Yoshino (1998); in this journal it has mostly been treated formally (Section 2), and modificatory provisions have been extracted from amending acts [Lesmo et al. 2012]. None of these grades an executable decision.

\textbf{Checking encodings against references.} Mlang passes the French tax administration's private, lawyer-written tests (all 476 usable cases for 2018; later 2018–2021) [Merigoux et al. 2021b]. Catala matched the official family-benefits simulator and found a rule omitted in OpenFisca [Merigoux et al. 2021a]; it also surfaced interpretive "micro-choices" in housing benefits through legal analysis and interviews rather than output comparison [Merigoux et al. 2023]. Tax-Calculator and PolicyEngine were compared with TAXSIM-35 for 2017–2021 and 2021–2025. Encoding exposes interpretive choices: legally trained coders of the Australian Copyright Act diverged unless they first agreed on key terms [Witt et al. 2021], a finding later developed into a validation methodology [Witt et al. 2023]; encoding has been argued to be an act of interpretation that open-textured standards resist [Yermukhametova et al. 2026]; competing readings of the Italian electoral law elect different deputies on the 2022 data, one reproducing the official outcome [Coppola 2026]; and a SAT solver finds the inputs on which LLM formalizations of EU provisions disagree [Vernie \& Grabmair 2026]. Catala has been presented in this journal as a basis for legal expert systems [Huttner \& Merigoux 2022]. LLM formalizations of US tax provisions [Yadamsuren et al. 2026] and LLMs on the VITA certification test [Gogani-Khiabani et al. 2025] are graded on one tax year against keys built by authors or experts; OpenExempt derives answers from expert encodings of US bankruptcy exemptions [Servantez et al. 2026]. In all of these the reference is the administration's own code, a calculator, the encodings themselves, or a single official outcome that confirms one reading; none grades executable rules against independent records of real outcomes across law versions, or uses such a record to test a reading of a contested provision.

\textbf{Evaluation.} LawShift perturbs statutes to test judgment prediction; TENFORTY applies metamorphic relations to tax software [Tizpaz-Niari et al. 2023]; TaxCalcBench grades direct LLM tax computation for 2024 against a commercial engine; the Axiom Foundation compares rules encoded with effective dates with TAXSIM and PolicyEngine, by default for one tax year [Axiom Foundation 2026].

\textbf{How the closest work is graded.} Table 3 uses only what each work reports.

\begin{table}[t]
\caption{What the closest work grades and against what.}
\footnotesize
\begin{tabularx}{\linewidth}{p{0.22\linewidth}XXXXX}
\toprule
Work & What is graded & Date of the law varies per case & Delegated or transitional provisions & Answer key & Years \\
\midrule
Law\&Order [Han et al. 2026] & executable tax program & no & no & independently authored returns (TaxCalcBench) & one \\
TaxCalcBench [Bock et al. 2025] & direct LLM tax computation & no & no & commercial tax engine & one (2024) \\
Synedrion [Gogani-Khiabani et al. 2026] & tax software & no & no Revenue Procedure selection & hand-written reference & one (2021) \\
Catala generation [Lorenzo et al. 2025; Ahmad et al. 2026] & Catala code & no & references, not versions in force & researcher-written & – \\
FiscalQA Pro, NitiBench+, LexKairos, Prior et al. & retrieved version or answer & yes & no & benchmark labels & several \\
TALD-Bench [Huang et al. 2026] & applicable law version & yes & transition to the Civil Code & court judgments & around the Civil Code \\
Axiom [Axiom Foundation 2026] & encoded rules & effective dates & – & TAXSIM, PolicyEngine & one by default \\
Mlang; Tax-Calculator, PolicyEngine [Merigoux et al. 2021b] & human-built tax code & several years, validated year by year & – & official test suite; TAXSIM & 2018–2021; 2017–2025 \\
Coppola 2026 & competing readings of an electoral law & no & – & official election outcome & one election \\
Vernie \& Grabmair 2026 & LLM formalizations of EU provisions & no & no & none (SAT-found disagreements, expert review) & – \\
LexTemporis (this paper) & executable status or amount rule & yes, each case as of its own date & yes & disclosures, TAXSIM, insurers' decisions & 9, 10, and 4 \\
\bottomrule
\end{tabularx}
\end{table}

\textbf{Coverage.} Of 61 studies we coded on temporal retrieval (A), statute-to-code (B), and delegation or structure-aware retrieval (C), 57 (93\%) test a single jurisdiction, and none combines A with B, or B with strict delegation C; the combination is what makes the phenomena in Section 6 visible.

\section{The test bed}

\subsection{Korea (main case)}

\textbf{Answer key.} Annual reports state whether the firm is an SME, in a table from FY2021 onwards and in prose before that; we extract the label with table rules and sentence classification (scripts/extract\_sme\_labels\_v3.py; rules in Appendix C). Version 3 of the extractor, written after an independent audit (Section 8), changed 219 labels and filled 355 previously unreadable ones. The disclosed status holds "as of the filing date of this report", in March of the following year, which falls in the application period that began on 1 April of the fiscal year (Decree Art. 3-3; Ministry of Government Legislation interpretation 14-0128). Of 11,572 labelled disclosures, 82 (0.7\%) cite a tax statute's SME definition; they stay in the evaluation, marked in the released labels (Appendix C). Excluding the 526 labels that rest on "not applicable" wording, agreement is 90.1\% instead of 89.1\%. The author checked a stratified sample of 200 disclosures by hand: of the 183 with a current label and a decidable text, 181 match (error 1.1\%, 95\% interval 0.3–3.9\%); both errors are status changes within the year. Eight labels coded non-SME from "not applicable" wording could not be decided by hand, whereas 8 of 9 unlabelled disclosures could (data/label\_check\_results.md).

\textbf{Facts.} Where possible, revenue and assets are taken from the earliest-filed annual report that contains each fiscal year, to limit leakage from later restatements (Appendix C). Revenue for FY2012 is missing, so the FY2015 decisions average two years instead of three (Decree Art. 7(2)). Implausible total-asset values, mostly unit errors, are treated as missing (271 rows; scripts/fin\_sanity.py; Appendix C); without this filter agreement is 88.7\%. Consolidated figures, used where reported, approximate the aggregation of subsidiaries' revenue under Annex 2 of the decree for majority-held subsidiaries; separate-entity figures lower agreement (Table 6). Each firm carries its current industry code for all years.

\textbf{Reference rule.} The reference rule is a hand-written implementation of the act, decree, Annex 1, the grace period, and the 2014 addendum. Because the preceding fiscal year's figures fix status in an application period (Decree Art. 3-3), figures above the ceilings change status only one period later. The reference rule starts the three-year grace period of Act Art. 2(3) with that period, so that SME status continues without a gap. This choice was fitted to the key and then checked on held-out firms (Section 6.2). Pre-2015 SME status, which decides whether the 2014 addendum applies, is taken from the firm's FY2014 disclosure, a source of the same kind as the key; the evaluated labels never enter the rule. The rule is one reading by the author and omits the ownership-based tests (Decree Art. 3, Annex 2; limits in Section 8). Its agreement with the key estimates how far the written rule, so read, determines the decisions; it does not measure the law's indeterminacy. Timeline examples assume fiscal years ending in December; for other year ends the application period moves with the fiscal year (Decree Art. 3-3(1)).

\textbf{Version store.} The store holds 39 versions of the act and 45 of the decree, with Annex 1 parsed into (KSIC range, ceiling) rows for each version.

\textbf{Dataset statistics.} The Korean key covers 10,901 firm-years of 1,677 listed firms with both a disclosed label and the facts for a decision (data/sme/kr\_numbers.md); 66\% are in manufacturing (industry shares in Appendix C). About 12\% of firm-years in every year have average revenue within ±20\% of their ceiling, where the decision is most sensitive to the rule.

\begin{table}[t]
\caption{Korean firm-years with a label and facts, by fiscal year.}
\footnotesize
\begin{tabularx}{\linewidth}{p{0.22\linewidth}XXXX}
\toprule
FY & Firm-years & Disclosed SME & Within ±20\% of ceiling & Label from prose / table \\
\midrule
2015 & 887 & 67\% & 13\% & 866 / 21 \\
2016 & 943 & 68\% & 13\% & 923 / 20 \\
2017 & 1,028 & 69\% & 13\% & 1,008 / 20 \\
2018 & 1,087 & 66\% & 12\% & 1,061 / 26 \\
2019 & 1,155 & 67\% & 12\% & 1,130 / 25 \\
2020 & 1,227 & 66\% & 14\% & 1,201 / 26 \\
2021 & 1,443 & 65\% & 13\% & 0 / 1,443 \\
2022 & 1,523 & 65\% & 11\% & 0 / 1,523 \\
2023 & 1,608 & 65\% & 12\% & 0 / 1,608 \\
All & 10,901 & 66\% & 12\% & 6,189 / 4,712 \\
\bottomrule
\end{tabularx}
\end{table}

The label source switches in FY2021 with the disclosure format; FY2015–2020 labels come mostly from sentence classification and carry more extraction risk; most of the audit's corrections fell in these years (Section 8).

\subsection{Comparison cases}

Two comparison cases use keys of other kinds: 2,000 synthetic US returns for tax years 2013–2022, graded by NBER TAXSIM35, and 7,845 German hospital site–procedure–year decisions on minimum volumes, graded by the insurers' decisions, with 24 versions of the minimum-volume rule (Mm-R). Their data, keys, conditions, and results are in Appendix F.

\section{Evaluating synthesized rules}

The components below, instruments of the evaluation rather than contributions, produce the LLM rules evaluated in Section 6.3 (component evaluation in Appendix A, procedure in Appendix C).

\subsection{Retrieval and the addenda reader}

For a decision date \emph{d}, the retriever takes the statute article in force on \emph{d}, the subordinate-rule articles in force on \emph{d} that cite it, articles one hop forward and backward, annexes they cite, and addenda alive on \emph{d}: those stating an end date on or after \emph{d}, or transitional or special-case rules at most \emph{L} years old (default \emph{L} = 5) that mention a retrieved provision. Addenda whose stated end dates have all passed are excluded. The synthesis runs of Section 6.3 used this retriever. The addenda reader, designed to replace its lifespan rule, splits each Korean addendum into articles and includes the substantive articles alive on \emph{d} that mention a provision already in the bundle, with no statute-specific keywords (Appendix C); it was evaluated as a component (Appendix A) but not used to build the synthesis inputs.

\subsection{Rule synthesis}

The bundle and a fixed interface specification go to the LLM, which returns one Python function \texttt{judge(f)} over the case facts, which runs with the Python standard library only. Korean conditions: no law; current law (one function for all years); date-correct fixed article list (Act Art. 2, Decree Art. 3, Annex 1); the same list padded with unrelated same-date provisions to our bundle's length (length control); our bundle; and an oracle bundle with exactly the required provisions (upper bound). The conditions of the comparison cases are in Appendix F.2. Each condition-year cell is sampled three times in separate conversations and reported as the mean and range over the three functions.

Before the remaining results were collected, the analysis plan fixed the primary comparisons (Korea: ours versus the date-correct fixed list; for the comparison cases, Appendix F.2), the primary subsets (Korean firms covered by the 2014 addendum in FY2015–2018 and all firms in FY2017–2019), and the success criteria; results seen before the plan was fixed are listed in it. A difference between two conditions is reported only if, in the primary subset, the mean over functions differs by more than the larger within-condition range and at least 7 of the 9 function pairs in a year point the same way. Results that miss these criteria are reported as no difference.

\textbf{Input schema.} The Korean interface includes a field stating whether the firm was an SME under the pre-2015 rules; the 2014 addendum needs it, and it also tells the model that an earlier regime existed. In the first isolated runs, a rule synthesized from current law without the addendum used the field to apply the three-year grace period of Act Art. 2(3) to such firms, reproducing the addendum's outcome until 2018 by a different legal route (the grace route). Under reading (b) of the addendum, the routes diverge afterwards, because that reading denies a later grace period (Section 6.2). All conditions receive the same schema.

\subsection{Verify-and-repair loop}

Properties are derived from the legal text by the author and fixed before the loop runs. They are the author's reading of the law, not an answer key; whether that reading is right is checked afterwards, once, against the independent key.

\begin{table}[t]
\caption{Properties used by the verify-and-repair loop (Korea).}
\footnotesize
\begin{tabularx}{\linewidth}{p{0.22\linewidth}XX}
\toprule
Case & Property & Source \\
\midrule
Korea & Equal decisions within one application period & Decree Art. 3-3 \\
Korea & Firms far below every ceiling are SMEs & Decree Art. 3, Annex 1 \\
Korea & Firms far above every ceiling with assets of KRW 500 bn or more are not & Decree Art. 3 \\
Korea & Firms covered by the 2014 addendum are SMEs in the 2016 period and not in the 2020 period & Addenda of 2014, Art. 2(1)–(2); reading (b) of (2), contested (Section 6.2) \\
Korea & Three-year grace after first exceeding & Act Art. 2(3) \\
\bottomrule
\end{tabularx}
\end{table}

The properties for the comparison cases are in Table F1. Inputs are synthetic; boundary dates that admit two readings (the end of the grace period, the addendum's deadline) are excluded and recorded as interpretive questions. A violating function is returned to the model, in the same isolated setting, with its code and up to five violating inputs and outputs but no answer-key value, for at most two rounds. Because the properties encode the author's reading, the loop's results depend on it; they are reported in Appendix D.5.

\subsection{Execution, isolation, and audit}

The main runs used Gemini 3.8 Flash through the agentic Antigravity command-line client on 8–9 November 2026. On 9 November, Claude Opus 5.5 and Claude Haiku 5.5 (Claude Code subagents) and GPT-OSS 120B (the same Antigravity runner) answered the same 343 prompts. A first round, run inside the project directory, was withdrawn after evaluation file names appeared in 33 of its 78 logs. On cases answered in both rounds, its no-law baseline had named the correct revenue ceiling in 100\% of cases, against 60\% in isolation (Appendix C). Every later run had its own folder outside the project, and every log or transcript was audited for project paths, answer-key file names, web use, and tool calls. For the Claude subagents isolation rests on instruction and audit, not on denied permissions (Appendix C).

\section{Results}

\subsection{RQ1: what the written rule explains and where the date enters}

\begin{table}[t]
\caption{Reference rule versus disclosed status (10,901 firm-years, 1,677 firms, FY2015–2023; data/sme/kr\_numbers.md).}
\footnotesize
\begin{tabularx}{\linewidth}{p{0.22\linewidth}XXX}
\toprule
Reference rule & Agreement & FY2015–17 & Within ±20\% of the ceiling \\
\midrule
Annex 1 and grace period, no addendum & 87.8\% & 83–86\% & 73.5\% \\
+ 2014 transitional addendum (reading (b) of Art. 2(2); reading (a): 88.5\%, Section 6.2) & 88.9\% & 89–90\% & 75.7\% \\
+ grace period once only (full reference rule) & 89.1\% & 89–90\% & 76.8\% \\
Separate instead of consolidated revenue (comparison; 11,079 firm-years) & 86.3\% & 87\% & 68.8\% \\
Full rule, grace counted one report later & 87.3\% & 89–90\% & 69.7\% \\
Full rule, pre-2015 status checked against FY2012–14 figures (35 FY2014 SME disclosures overruled) & 88.8\% & 87–88\% & 76.8\% \\
\bottomrule
\end{tabularx}
\end{table}

\textbf{Uncertainty.} With firms resampled (2,000 bootstrap draws, data/robust\_ci.md), agreement with the disclosures is 0.891 (95\% interval 0.879–0.903). Modelling the 2014 addendum together with the once-only rule (the full rule compared with the first row of Table 6) raises agreement by 0.013 (0.010–0.018) over all years and by 0.051 (0.037–0.067) in FY2015–2017, when the addendum was in force; the addendum alone accounts for 1.1 points.

\textbf{Amounts versus transitional provisions.} Applying the 2025 annex to FY2015–2023 flips 3.5\% of size decisions (407 of 11,691) but changes the size test's agreement with the disclosures only from 85.3\% to 85.8\% (data/sme/metamorphic.txt). The 2014 addendum has a larger effect. Of 98 firms that failed the size test in a FY2015–2017 report yet disclosed SME status in all three years, 27 of the 94 with a FY2018 disclosure became non-SMEs in FY2018 (29\%, against 5.5\% of all firms that were SMEs in FY2017). Status-fixing rules have a similar effect (Appendix D.4). In Korea, the date matters through transitional and status-fixing rules more than through the amounts.

\textbf{What remains.} We drew a stratified sample of 60 of the remaining disagreements (11\%, 1,183 firm-years), weighted to current disagreements. In this sample, about 32\% point to the independence requirements that the rule omits, and 27\% to a grace period used before our data begin (Appendix D.1). The author classified the sample without ownership data or certificates; in it, most of the gap fits missing facts better than misread law. Membership in a business group designated by the Korea Fair Trade Commission (KFTC) alone accounts for 10.5\% of the disagreements.

\textbf{Business-group sensitivity} (added after the main results; scripts/group\_sensitivity.py, data/sme/group\_sensitivity.md; KFTC Open API data on designated business groups, data ID 15091891, downloaded 2026-11-14; raw data not released under the API's terms). Under the designation for their fiscal year, 333 of the 10,901 firm-years (76 firms) belong to a designated group. Of these, 12 disclose SME status and 321 non-SME status, whereas the reference rule, which omits the independence test, classifies 136 as SME and 197 as non-SME. These firm-years account for 124 of the 1,183 disagreements.

\begin{center}
\footnotesize
\begin{tabularx}{\linewidth}{p{0.22\linewidth}XXXX}
\toprule
Sample & Firm-years & Agreement & Disagreements & 2014 addendum readings where they differ: (a) / (b) \\
\midrule
All & 10,901 & 0.891 & 1,183 & 59 / 130 \\
Without designated-group firm-years & 10,568 & 0.900 & 1,059 & 59 / 128 \\
\bottomrule
\end{tabularx}
\end{center}

Excluding these firm-years changes the sample; it does not implement the independence test (Section 8).

\textbf{The comparison cases.} In the comparison cases, the date enters through other layers (Appendix F.3). In US income tax, it enters through the amounts in the yearly delegated Revenue Procedures, so that the latest document before a date governs the wrong year. In German minimum volumes, it enters through transitional clauses and an evaluative review standard whose application changed from year to year.

\subsection{RQ1: the keys test the author's reading}

\textbf{Grace-start convention.} The key also tested the author's choice of when the grace period starts. Act Art. 2(3) grants the grace period "for three years from the year following the year in which the cause arose", and status in an application period is fixed by the preceding fiscal year's figures (Decree Art. 3-3, Art. 7). Consider a December-year firm whose three-year average revenue first exceeds its ceiling with the FY2018 figures:

\begin{center}
\footnotesize
\begin{tabularx}{\linewidth}{p{0.22\linewidth}XXXX}
\toprule
Report (filed) & Application period containing the filing date & Figures that fix status & Reference rule (earlier start) & Later start \\
\midrule
FY2018 (Mar 2019) & Apr 2018 – Mar 2019 & FY2015–2017 average, below ceiling & SME (size) & SME (size) \\
FY2019 (Mar 2020) & Apr 2019 – Mar 2020 & FY2016–2018 average, above ceiling & SME (grace year 1, 2019) & non-SME \\
FY2020 (Mar 2021) & Apr 2020 – Mar 2021 & above & SME (grace year 2, 2020) & SME (grace) \\
FY2021 (Mar 2022) & Apr 2021 – Mar 2022 & above & SME (grace year 3, 2021) & SME (grace) \\
FY2022 (Mar 2023) & Apr 2022 – Mar 2023 & above & non-SME & SME (grace) \\
\bottomrule
\end{tabularx}
\end{center}

The grace years are 2019–2021. The reference rule gives three grace periods and no gap; the later start leaves the firm without SME status for one period and then grants grace. The reference rule's reading seems the more natural one, although "year" can also be read as the period in which status is lost, which gives the later start. We chose it, however, because of the disclosures. Among firms to which the 2014 addendum never applied (so that the comparison does not depend on the addendum's two readings; see below), the two conventions differ on 370 labelled firm-years, and the disclosures side with the earlier start in 280. We made the change after seeing this comparison over all years and recorded it in the analysis plan before any LLM output was scored. It raises agreement from 87.3\% (later start) to 89.1\%.

\textbf{Split-sample check of the grace start.} Because the start was chosen on the key, we checked the choice on held-out firms (scripts/grace\_split.py; data/sme/grace\_split.md). Firms are split into halves. The start is chosen on one half, using the firm-years outside the 2014 addendum where the two starts differ, and is evaluated on the other half. In the split fixed before the run (seed 20261114), the choice half picks the earlier start (116 of 165). On the held-out half, this start agrees with the disclosures on 0.892 of firm-years, against 0.868 for the later start, and the held-out firm-years on which the addendum's readings diverge side with (a) in 37 and with (b) in 70. Over 1,000 random splits, the choice half picks the earlier start in 1,000. The median held-out gain in agreement from the chosen start is +0.018 (95\% range +0.015 to +0.022), and the median held-out share of diverging addendum firm-years that favour (b) is 0.683 (0.586 to 0.789). Under the later start, the split of the readings on the whole sample is still 55 (a) to 118 (b). The RQ1 results thus do not rest on a convention fitted to the key alone; the period and firm splits are in Appendix D.4.

\textbf{An interpretive question exposed by the key.} Article 2(2) of the 2014 addendum reads: "firms deemed SMEs under paragraph 1 shall not be deemed SMEs under Article 2(3) of the Act" (the grace period; in Korean, "제1항에 따라 중소기업으로 보게 된 기업에 대해서는 법 제2조제3항 본문에 따른 중소기업으로 보지 아니한다"). It admits two readings, which diverge on 1 April 2018, the day after the deeming ends:

\begin{itemize}
\item (a) a deemed firm cannot hold the grace period \emph{while} deemed; when the deeming ends, it loses SME status and then receives the three-year grace period, so a firm still above the ceilings remains an SME in the FY2018–FY2020 reports;
\item (b) a deemed firm never receives the grace period, even after the deeming ends, so the same firm is a non-SME from the FY2018 report onwards.
\end{itemize}

The parallel transitional provision in the 2015 addenda to the Act has no such sentence. We found no official interpretation, ministry guide, published ministry reply, or court decision that addresses the grace period after the deeming ends, and the reviewing lawyer knew of none (Appendix D.4; Section 8). The reference rule (Section 4.1) follows (b).

Under reading (a), following the convention the reference rule uses for the ordinary grace period, grace starts with the FY2018 report (the year the deemed status ended) and runs through FY2020. Overall agreement is then 88.5\%, against 89.1\% under (b). The readings differ on 189 labelled firm-years of 78 firms; the disclosures side with (a) in 59 and with (b) in 130 (0.69; 95\% interval 0.59–0.78 with firms resampled). Counted by firm, 51 firms side mostly with (b), 26 mostly with (a), and one is tied (exact sign test on the 77 untied firms, p = 0.006; data/robust\_ci.md). The author checked all 189 disclosures by hand against the report texts. Every label that could be decided matched; the 19 that state only "not applicable" could not be decided. Without them, the split is 59 to 111 (by firm, 44 to 26, p = 0.04; data/label\_check\_results.md), and 60 to 110 according to a blind second annotator (Section 8). The lean to (b) holds when pre-2015 status is checked against FY2012–2014 figures (Table 6), when grace is counted one report later, when coverage is restricted, and without the "not applicable" labels. Before the audit's label corrections, the split was close (Appendix D.4).

The disclosure text points the same way (Appendix D.3). The six firms that state when their status ended all fit (b), and no firm states (a). The firms whose labels fit (a) never link a grace period or the addendum to their status, and nearly all of them repeat, word for word, the sentence they wrote while the deeming applied. This fits reporting inertia better than a deliberate reading.

\textbf{Interpretive canons and legal review.} Read by the usual methods of interpretation, the provision points to (b) but does not settle the question. \emph{Literal:} Paragraph 2 names the firm, not a period; it lacks words such as "while it is deemed an SME", which reading (a) would need. \emph{Systematic:} where the drafters wanted a further deemed period, they said so (Article 3 of the same addenda: "once only"). The transitional addendum of the 2009 decree amendment (Presidential Decree No. 21368, Addenda Art. 2), which kept firms SMEs until 31 December 2011, had no sentence on the grace period, so the 2014 drafters added Paragraph 2 to a model that lacked it. Against this, the 2015 addenda to the Act have no parallel sentence. \emph{Teleological and historical:} the stated reasons for the 2014 amendment include limiting to once the number of times a firm that outgrows the SME criteria is deemed an SME (item d(2); Decree Art. 9). Reading (a), by contrast, would extend the protection of covered firms from about three years and three months (1 January 2015 to 31 March 2018) to about six years and three months (to 31 March 2021). The transitional deeming protects the reliance of firms classified under the old criteria, and (b) serves that purpose fully. Since the boundary of a transitional provision is the legislature's choice, and expectations beyond it are not protected [Constitutional Court 2019], nothing in constitutional doctrine requires reading a further grace period into Paragraph 2. \emph{Strict construction:} a deeming provision attaches a status by legal fiction. The Framework Act is not a tax act, but SME status also matters in tax law, where the Constitutional Court has held that a lapsed provision cannot be extended by interpretation beyond what the legislature enacted [Constitutional Court 2012]; by analogy, reading a further fiction into the period after the deeming, as (a) does, goes beyond the words. Paragraph 2 still has a function under (a): it prevents figures above the new ceilings during the deeming from starting a grace period alongside the deeming. The canons converge on (b), but none is decisive, and no court or ministry has decided the question. A practising lawyer also reads Paragraph 2 as (b) and recommends presenting both readings (Section 8). The disclosures add what the canons cannot supply: how the regulated firms applied the provision. We therefore keep (b) in the reference rule and report results under both readings.

\textbf{Why the disclosures bear on meaning.} The disclosures are not an authority on what the provision means; a court would decide that. Nor do they make a firm an SME. The Ministry of Government Legislation reads "deemed" in the grace provision as a legal fiction, with status following the statutory requirements rather than the firm's will, so a firm cannot waive it (interpretation 15-0157, 2015); the 2014 addendum uses the same word. Two features give the disclosures weight. First, a non-SME statement goes against the firm's interest, since SME status carries tax, procurement, and financing benefits. Second, because the status is not at the firm's disposal, such a statement is more plausibly a considered application of the rule than a choice. The (a)-side disclosures mostly repeat the earlier sentence (Appendix D.3), so the non-SME disclosures on the (b) side carry more weight than their count alone suggests. Filed under a disclosure duty and relied on by regulators and lenders, the disclosures are evidence of how the rule's addressees applied it. Where no official interpretation exists, that practice is the best available record of how the rule operated, and it is the record against which an executable rule will be compared in use. A reading that the practice contradicts needs a reason; a reading that the practice follows still needs a legal argument, hence the canons above.

\begin{figure}[t]
\centering
\includegraphics[width=\linewidth]{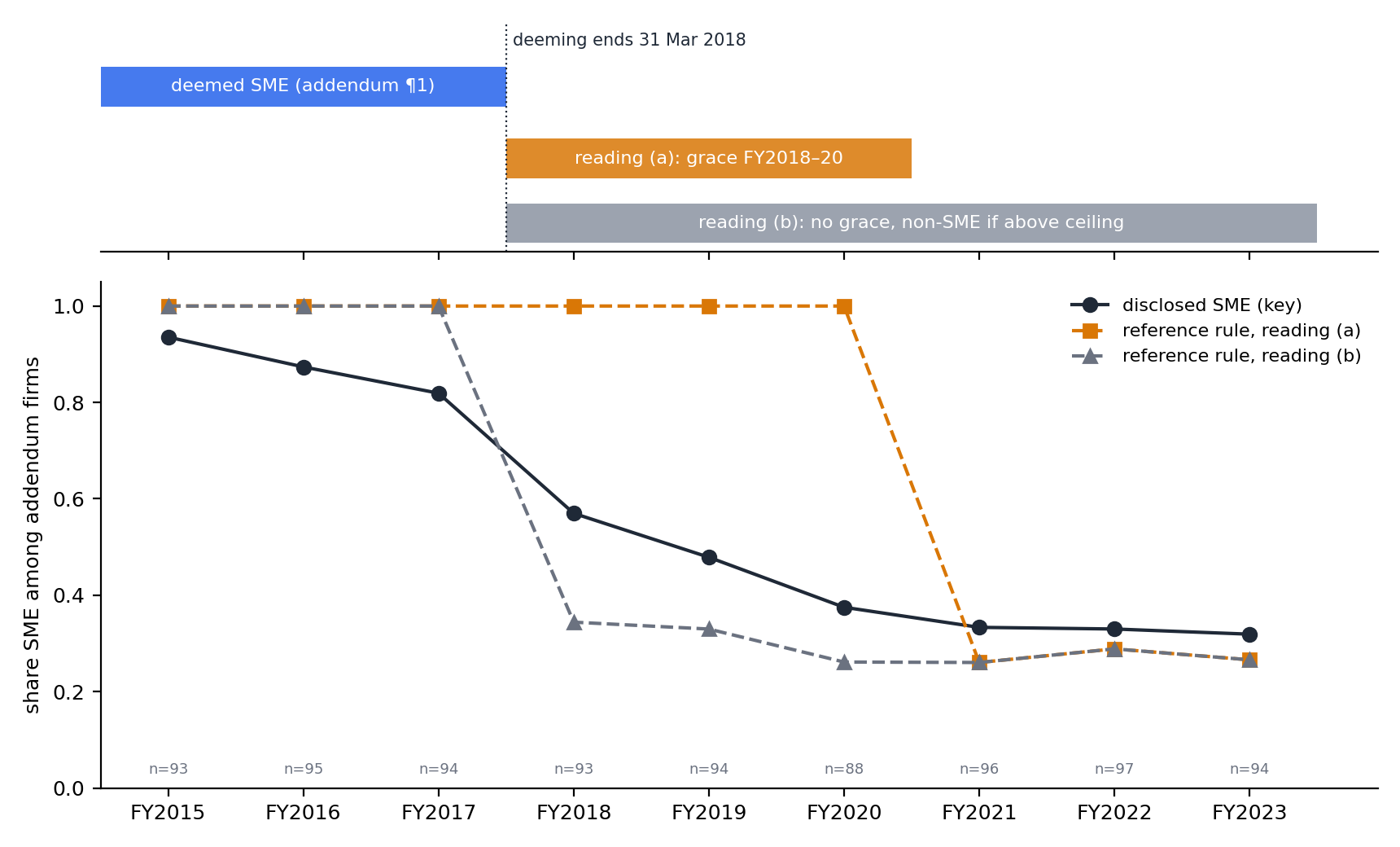}
\caption{Firms covered by the 2014 addendum. Top: the deeming period and the two readings of Paragraph 2. Bottom: share disclosing SME status (key) and share predicted to be SMEs by the reference rule under each reading. The disclosures fall gradually from FY2017 to FY2021 and follow the step of neither reading.}
\end{figure}

An independent key cannot decide which reading is right; here it showed that the disclosures lean to (b) while some remain consistent with (a). The repair property for the 2014 addendum (Table 5) encodes reading (b), so the repair loop's results inherit this question (Appendix D.5).

\subsection{RQ2: grading synthesized rules against two references}

The pre-registered comparison of retrieval conditions finds no condition that is better across the four models. The primary comparison sets our bundle against the date-correct fixed list, per year, on two subsets: subset A (firms covered by the 2014 addendum, FY2015–2018) and subset B (all firms, FY2017–2019). No cell of this comparison passes in three or more models; eight of fourteen pass in one or two (Claude Opus 5.5 and Claude Haiku 5.5), and six pass in none (Table D1; analysis plan, change log of 9 November 2026). For Gemini 3.8 Flash, the main model, no year of either subset passes against either key. The verdicts differ by model because models differ in whether they already know, or route around, the provisions that a bundle supplies (the grace route, Section 5.2). The results cover all 303 synthesis outputs after the regeneration of invalid responses (Appendix C). Verdicts under seven sensitivity settings, the controls, and the per-model analyses are in Appendix D.2. Table 7 gives the Gemini results for every Korean condition.

\begin{table}[t]
\caption{Korean conditions including the controls (mean over functions [range]; n = functions). Subset A: firms covered by the 2014 addendum, FY2015–2018, as pre-registered.}
\footnotesize
\begin{tabularx}{\linewidth}{p{0.22\linewidth}XXXX}
\toprule
Condition & All, vs reference rule & All, vs disclosures & Subset A, vs reference rule & Subset A, vs disclosures \\
\midrule
Exact provisions (oracle) & 0.989 (27) & 0.890 & 0.922 [0.735–1.000] (12) & 0.822 \\
Ours & 0.989 (27) & 0.890 & 0.895 [0.714–1.000] (12) & 0.817 \\
Date-correct fixed list & 0.968 (27) & 0.884 & 0.851 [0.327–1.000] (12) & 0.777 \\
Fixed list padded to our size & 0.967 (27) & 0.884 & 0.776 [0.265–1.000] (12) & 0.745 \\
No law & 0.939 (25) & 0.866 & 0.879 [0.622–1.000] (12) & 0.801 \\
Current law & 0.945 (3) & 0.863 & 0.875 [0.831–0.898] (3) & 0.805 \\
\bottomrule
\end{tabularx}
\end{table}

The rest of this section is exploratory. Criteria for a confirmatory re-test on new models (hypotheses H1 and H2 and the rules for failure) were fixed in the analysis plan's change log of 14 November 2026, before any new output; no such re-test has been run.

\textbf{Shrinkage against the key.} Pooled over years, with firms and (within each condition and year) functions resampled (data/robust\_ci.md), our functions lead the fixed list by 0.021 (95\% interval 0.013–0.029) against the reference rule but only by 0.006 (0.002–0.011) against the disclosures. They lead the no-law condition by 0.050 (0.040–0.061) and 0.024 (0.016–0.033), respectively. Over all firm-years, our bundle's lead over each other condition is 28–60\% as large against the disclosures as against the reference rule (data/robust\_null.md). The lead shrinks in the same way for Opus and Haiku, but not for GPT-OSS 120B, whose functions are far from both keys (Appendix D.2). Part of the shrinkage arises by construction: where the reference rule is wrong, a function that reproduces it more faithfully must lose against the key. A null ratio of 2p − 1 = 0.78 (p = 0.891, the reference rule's agreement with the disclosures) gives the shrinkage expected if functions departed from the reference rule independently of whether it is right. This null was fixed after the Gemini and Opus results had been seen. All four Gemini ratios lie below it, including the upper ends of their 95\% intervals (Appendix D.2).

\textbf{Is the concentration only difficulty?} The firm-years where the reference rule is wrong are also the harder ones: they are more often near the ceiling or covered by the 2014 addendum. We formed four strata of difficulty (average revenue within ±20\% of the ceiling or not, crossed with coverage by the 2014 addendum). Within them, the lead of our bundle against the reference rule is still larger where the reference rule disagrees with the disclosures than where it agrees. The stratum-weighted difference (disagree minus agree, with a 95\% firm-cluster bootstrap) is +0.067 [+0.040, +0.097] against no law, +0.032 [+0.008, +0.059] against current law, +0.052 [+0.036, +0.069] against the fixed list, and +0.049 [+0.034, +0.067] against the padded list (Gemini; scripts/robust\_strat.py, data/robust\_strat.md). Difficulty alone does not explain where the lead lies: part of what a fuller bundle adds against the reference rule is a more faithful reproduction of the author's reading, including where that reading is wrong.

\textbf{Where retrieval helps against the key.} Where a model does not reach the addendum's outcome by the grace route, the provision that the bundle adds decides the outcome, and the lead survives the independent key (Opus, subset A, FY2015: 0.935 versus 0.348, measured against the disclosures). In the comparison cases, the model already knows the law, and the pre-registered comparison supports no claim for our bundle; models without the amounts in memory, however, gain from retrieval (Appendix F.4). Together, these results suggest a hypothesis for further models, not a finding: retrieval matters where the model does not already know the law, and those gains survive the independent key.

\subsection{What the synthesized rules got right and wrong}

Two functions show the patterns behind the numbers; two US examples are in Appendix F.5.

\emph{A different legal route to the same outcome (Korea, current law).} The function synthesized from current law has no addendum to read, but it uses the input field on pre-2015 status to start a grace period:

\begin{verbatim}
pre2015 = f.get("pre2015_sme")
if start_year <= 2015:
    prev_is_sme = True if pre2015 is True else False
\end{verbatim}

It treats a firm that exceeds the 2015 ceilings as leaving SME status and grants the three-year grace period of Act Art. 2(3). Through FY2017, this gives the addendum's result by another provision; from FY2018, the routes differ (Sections 5.2 and 6.3).

\emph{Reasoning about the grader (Germany).} The one German function that returned no value is not a reversed rule. Its response contains working notes that speculate about "the benchmark author's reference code" and place a fragment of that guessed code in the first code block. Our scorer, written before the runs, takes this first block; the complete function follows later. All 14 Korean responses whose first block did not load also contain two to nine code blocks. After the regeneration (Appendix C), taking the last block instead (a post-hoc sensitivity analysis) changes no verdict (data/synth/criterion\_\_lastblock.md; Appendix D.2). An agentic model that suspects a hidden reference implementation may try to match it rather than the law. This is one more reason to keep keys and reference rules out of reach (Section 5.4).

\section{Discussion}

\textbf{Three ways the date matters.} In Korea, transitional and status-fixing rules decide a visible share of cases, while the 2025 change of annex amounts barely matters. The comparison cases show two other ways (Appendix F.3). In the US, the amounts change every year, and the delegated document must match the tax year, not the date. In Germany, the applicable value sits in a transitional clause, and the decision follows an evaluative standard whose application the rule changed year by year. Selecting "the version in force on the date" handles none of these on its own.

\textbf{Relation to formal models of legal time.} Table 2 maps the three cases onto the times of the norm that formal models of legal change distinguish [Tamargo et al. 2019; Moguillansky et al. 2023] (Section 2). In Korea, tracking only the time in force misses provisions that move a visible share of decisions.

\textbf{Interpretation.} Disagreements with an independent key mix missing facts, reporting errors, and genuine interpretive questions. The provisions where the Korean reference rule is weakest are exactly those on which the Ministry of Government Legislation has issued interpretations. The 2014 addendum is the opposite case: the disclosures, including the six firms that state when their status ended (Appendix D.3), and a practising lawyer lean to reading (b), but no authoritative interpretation exists (Section 6.2). A synthesizer graded against a single reading would be scored right or wrong on an open legal question. Adding dated interpretations as a further retrieval layer under the same as-of-date filter is left to future work. Linking ownership data, such as the KFTC's business-group designations and the affiliate disclosures in the same annual reports, would let a rule apply the independence requirements behind the largest group of disagreements (Section 6.1). Pairing each case with the versions before and after an amendment would directly measure the bias towards the newer law that TALD-Bench reports. Here, the current-law condition shows this bias only where the current text no longer carries the old rule (Korea), not where it does (Germany; Appendix F.4).

\section{Threats to validity}

This section collects the study's limitations in one place; the results sections state each finding without repeating them.

\textbf{The key is the firms' self-assessment, not a legal ground truth.} Each key is independent of the author's reading, which is what the evaluation needs, but none is ground truth in a legal sense. Disclosed status is the firm's own judgment, TAXSIM is a research calculator maintained by others, and the German decisions include self-reported counts and evaluative judgments. The determinacy analysis is thus in part an analysis of each key. The Korean disclosures are self-assessments outside the audit of the financial statements (a firm may misread the law, repeat an earlier statement, apply another statute's definition, state a status without a certificate, or conservatively decline SME status). Part of the 11\% disagreement may therefore be error in the disclosures, which cannot be separated from error in the rule without the certificates. TAXSIM approximates the personal-exemption phase-out, reports AMT on a separate line, and does not cover 2023 (Appendix F.1), and the German decisions follow an evaluative review standard that no threshold rule can reproduce (Appendix F.3). We report these as properties of the keys rather than correcting them.

\textbf{Errors in the key; the label corrections changed one result.} An independent audit of the first Korean label extractor found about 2\% of evaluated labels wrong, including FY2014 labels that decide the 2014 addendum, so the errors fell where the key decides the addendum question; version 3 of the extractor corrects them (Section 4.1). Agreement was 89.2\% on 10,601 firm-years with the earlier labels and 89.1\% on 10,901 with the corrected ones. However, the corrections turned a close split between the readings of the 2014 addendum into the lean to (b) (Appendix D.4) and reversed the direction of the repair loop's result (Appendix D.5). Label errors therefore matter where few cases decide a question. The author's hand check of 200 stratified labels found 2 errors in 183 (Section 4.1) and confirmed every decidable label of the 189 firm-years on which the readings of the 2014 addendum differ (Section 6.2). A second annotator, an undergraduate not involved in the project and blind to the labels and to the author's decisions, judged all 385 disclosures with the same instructions and agrees with the author on 352 of the 353 that both decided (Cohen's κ = 0.99; data/label\_check\_human2.md). Over all three categories, agreement is 91\% (κ = 0.84). The difference arises because the second annotator coded 28 disclosures that say only "not applicable" as non-SME, as the extractor does, although the instructions asked for "undecided"; the author left them undecided. On the 170 addendum firm-years that the author could decide, the second annotator's split is 60 to 110.

\textbf{One non-lawyer author's reading.} The reference rules are one reading by an author who is not legally trained, and the 2014 addendum showed that such a reading can be contested (Section 6.2). One practising lawyer reviewed the four readings on which the Korean reference rule rests, agreed with the timing of the application period and the start of the grace period, and read the 2014 addendum as (b). One reviewer cannot settle a contested provision, so a difference that holds against the reference rule but not against the key is reported as a property of the reading. Every source cited in Sections 2 and 6.2 and Appendix B.2 was checked in the National Law Information Center, Crossref, KCI, or the original publication; an AI-assisted legal analysis was used only to find leads.

\textbf{The grace start was fitted to the key.} The convention for the start of the grace period was chosen after comparing both starts with the disclosures over all years (Section 6.2). The split-sample check on held-out firms confirms the choice and the lean to reading (b). However, its halves come from the same disclosures, so it guards against fitting to particular firms, not against a bias shared by the whole key.

\textbf{The independence test is omitted.} The Korean reference rule omits the independence requirements (Decree Art. 3(1)(ii), Annex 2). The reviewing lawyer judged that this matters for listed firms, which more often belong to large business groups or have related firms whose figures are combined. The independence requirements are also the largest group in the sample of remaining disagreements (Section 6.1; Appendix D.1). A sensitivity analysis excluding firm-years of KFTC-designated business groups, added after the main results, raises agreement from 0.891 to 0.900 and leaves the split of the addendum readings at 59 to 128 (Section 6.1). Designated-group membership alone accounts for 10.5\% of disagreements, against the sample estimate of about a third for the independence requirements as a whole. This analysis changes the sample rather than implementing the independence test: affiliates of undesignated groups and the 30\%-ownership and affiliated-firm aggregation rules remain unmodelled.

\textbf{RQ2 is exploratory, with three functions per cell.} The LLM results come from four models with three samples per cell, which are often nearly identical within a cell. Only the pre-registered comparison is confirmatory, and its verdicts differ by model, because models differ in whether they already know, or route around, the provisions that a bundle supplies; which retrieval condition is better is therefore model-specific. The shrinkage analyses of Section 6.3 are exploratory: the null ratio was fixed after the Gemini and Opus results had been seen, and the stratified re-analysis was added afterwards. The criteria for a confirmatory re-test are fixed in the analysis plan (change log of 14 November 2026). The isolation of the Claude subagents rests on instructions and an audit of every transcript rather than on denied permissions (Section 5.4), and the author also used Claude as an assistant (Declarations).

\textbf{Changes after pre-registration.} The analysis plan (repository document 18) lists every change made after the plan was fixed: the grace-start convention (before any LLM output was scored); the post-hoc FY2019–2020 subset; the code-block sensitivity; after the audit of 9 November, label extractor version 3, the financial-data filter, regeneration of the 16 graded responses whose first code block did not load (the 17th, US 2023, lies outside the graded years), and the correction of subset A of Table 7 to the pre-registered definition; reporting every result against both the reference rule and the key; and, on 14 November 2026, the criteria for a confirmatory re-test, with the grace start confirmed by the split-sample check. Primary results use the rules as fixed.

\textbf{Required-provision lists and gold labels.} The author wrote the retrieval lists in the same session as the retriever, and wrote the addenda reader's gold labels after the reader was built; no independent human annotation is yet available (Appendix A). A blind second annotation by an LLM annotator agrees on relevance for 25 of 29 articles (Cohen's κ = 0.73) and on the end date for 5 of the 8 articles for which either annotator gives one. All four disagreements are articles whose effect ended before 2015, which the annotator marked irrelevant, as instructed. On whether an article concerns the status at all, the two therefore agree throughout (data/addenda\_second\_annotation.md). The reader's transfer was checked on only one further statute, the Mid-sized Enterprise Act.

\textbf{Scope.} Korean listed firms only; synthetic US taxpayers with a narrow income profile; German data for four entitlement years. Besides the independence requirements (above), the Korean reference rule omits the exclusion from the grace period after a merger with a firm in its grace period (Decree Art. 9(i)), the annualized revenue of firms founded or merged within the preceding year (Decree Art. 7(2)(iii)), determination of the main industry beyond the reported code, and changes of fiscal year. The FY2015 decisions average two years of revenue instead of three. The 2024 amendment extending the grace period to five years (Act Art. 2(3), amended 20 February 2024) falls after the study period.

\section{Conclusion}

Executable legal rules are usually checked against the reading of the person who wrote them. LexTemporis checks them against keys written by others, chiefly what Korean listed firms disclosed about their own SME status. It releases the test bed, version stores, and scoring scripts so that other systems, models, and readings can be measured in the same way. Graded this way, the author's own rule shows how far the written rule determines decisions and where the date enters: through a transitional addendum and status-fixing provisions more than through the annex amounts. The disclosures also test the author's reading. They favour reading (b) of the contested 2014 addendum in 130 of 189 diverging firm-years. This lean appeared only once errors in the extracted labels were corrected, and hand checks and a blind second annotator confirm it. The grace-start convention fitted to the disclosures also holds on held-out firms. The comparison cases (Appendix F) show the date entering through yearly delegated amounts in US income tax and through transitional clauses and an evaluative standard in German minimum volumes; in US tax, the synthesized rules follow the statute more closely than the key does. For LLM synthesis, the pre-registered comparison finds no retrieval condition that is better across four models. That the lead of our retrieval shrinks against the key is an exploratory observation, partly by construction; a confirmatory test is specified but has not yet been run. The reading of the law is part of what has to be evaluated, and an answer key written by someone else makes that possible.

\begin{appendices}
\counterwithin*{table}{section}\counterwithin*{figure}{section}
\renewcommand{\thetable}{\thesection\arabic{table}}\renewcommand{\thefigure}{\thesection\arabic{figure}}
\section{Evaluation of the retrieval components}

The retriever (Section 5.1) and the addenda reader are components of the baseline system. This appendix evaluates them against the author's gold labels, which, unlike the answer keys, are not independent of the author (Section 8).

\textbf{Provenance of the required-provision lists.} The list of provisions a Korean SME decision requires (Act Art. 2; Decree Arts. 3, 3-3, 7, 9; Annex 1; the 2014 addendum until 2018-03-31; the 2025 addendum from 2025-09-01) was written by the author in the same session as the retriever, not fixed in advance. We therefore report the retriever's sensitivity to its single parameter, a size-matched baseline, and its transfer to a statute it was not built on.

\textbf{Two versions.} The retriever of Section 5.1 includes whole addenda by a lifespan heuristic; the synthesis runs used this version. The addenda reader of Section 5.1 is designed to replace the heuristic. Table A2 scores both against the original required-provision list; the revised list is reported below.

Gold labels for the addenda reader cover the 24 substantive articles of the SME act and decree and record two levels of relevance (within the experimental specification; legally relevant including independence and merger rules), start and end dates, and the evidence: 7 articles have explicit end dates that can be checked mechanically, and the rest are labelled from the text with a recorded rationale. The labels were written after the reader was built; we therefore also apply the reader unchanged to one further statute, the Mid-sized Enterprise Act, whose addenda contain one relevant article. This checks transfer on one statute only.

\begin{table}[t]
\caption{Addenda selection at article level, 1 April of each year 2015–2026, legally relevant gold.}
\footnotesize
\begin{tabularx}{\linewidth}{p{0.22\linewidth}XXXX}
\toprule
Statute & Method & Precision & Recall & Articles included \\
\midrule
SME act and decree & Five-year lifespan heuristic & 0.30 & 0.52 & 143 \\
SME act and decree & Addenda reader & 1.00 & 1.00 & 83 \\
Mid-sized Enterprise Act and decree & Five-year lifespan heuristic & 0.12 & 0.67 & 48 \\
Mid-sized Enterprise Act and decree & Addenda reader & 1.00 & 1.00 & 9 \\
\bottomrule
\end{tabularx}
\end{table}

The reader reads the end date of all 24 gold articles correctly and the start date of 22. The heuristic's recall of 0.52 shows that the retriever's recall of 1.00 (Table A2) depended on a required-provision list with only two addenda. With a revised list that adds Decree Art. 7-2 (asset calculation, required by the specification) and every addendum article marked relevant to the specification, the five-year heuristic reaches recall 0.91 (3 of 12 years complete) and the reader 1.00 (12 of 12). Bundle precision barely moves (0.36 to 0.37), because one-hop expansion over decree articles, not addenda, dominates bundle size. The revised list was written after inspecting results and is reported alongside the original.

\begin{table}[t]
\caption{Korean SME act, decision date 1 April of each year, 2015–2026, original required-provision list (above). Recall and precision count required provisions at the correct version; precision is strict. Against the revised list (above), the lifespan heuristic reaches recall 0.91 and the addenda reader 1.00. Bundle size counts items: the lifespan heuristic adds each addendum as one item, the addenda reader adds individual addendum articles, so its bundles hold less addendum text but more items.}
\footnotesize
\begin{tabularx}{\linewidth}{p{0.22\linewidth}XXXXX}
\toprule
Method & Recall & Precision & F1 & Avg. bundle size & Years with recall 1.0 \\
\midrule
Current law, fixed list & 0.38 & 0.33 & 0.35 & 7.0 & 0/12 \\
Date-filtered statute only & 0.16 & 1.00 & 0.27 & 1.0 & 0/12 \\
Date-correct fixed list & 0.48 & 1.00 & 0.64 & 3.0 & 0/12 \\
BM25, same size as ours & 0.68 & 0.20 & 0.31 & 22.0 & 0/12 \\
\textbf{Ours (L = 5)} & \textbf{1.00} & 0.29 & 0.45 & 22.0 & \textbf{12/12} \\
Ours with addenda reader & 1.00 & 0.27 & 0.42 & 23.9 & 12/12 \\
\bottomrule
\end{tabularx}
\end{table}

\textbf{Sensitivity.} With only explicit end dates (\emph{L} = 0) recall is 0.99 (11/12 years); with \emph{L} = 3, 4, 7, 10, or no limit it stays 1.00, and bundles grow from 17.8 to 25.3 provisions. Against the original list, the result does not depend on fitting \emph{L} to the 2014 addendum; that list contains only two addenda, and Table A1 shows that the heuristic misses other relevant addenda.

\textbf{Precision.} Bundles are large. The retriever is recall-oriented by design, and selection within the bundle is left to synthesis; a fixed list that already knows which articles matter has higher F1, but cannot be built for a new statute without the knowledge the task asks for.

\textbf{Transfer.} Applied unchanged to the Mid-sized Enterprise Act (Art. 2 → Decree Art. 2, amended 2016, 2018, 2021), the retriever recovers both required provisions at the correct version in all 12 years (recall 1.00, 11.5 provisions per bundle); the same walk over current versions recovers 0.25 and date-only retrieval 0.50. The walk also surfaced an addendum, deleted by a 2023 amendment, that had limited the act's validity to ten years; addenda are themselves amended and must be versioned.

\textbf{US analogue.} Selecting the Revenue Procedure that governs tax year \emph{Y} (Table A3).

\begin{table}[t]
\caption{Revenue Procedure selection, tax years 2013–2023.}
\footnotesize
\begin{tabularx}{\linewidth}{p{0.22\linewidth}XX}
\toprule
Method & Correct at filing date (Y+1-04-15) & Correct at year end \\
\midrule
Latest overall & 0/11 & 0/11 \\
Latest issued before the date & 0/11 & 0/11 \\
Procedure number year = Y & 2/11 & 2/11 \\
\textbf{Applicability read from text} ("taxable years beginning in Y"; latest if superseded) & \textbf{11/11} & \textbf{11/11} \\
\bottomrule
\end{tabularx}
\end{table}

Revenue Procedures are issued in the autumn for the following year, so the most recent document before a date governs the wrong year. In 2018 the document in force and the applicable document also diverge: on 1 February 2018 the 2018 amounts in force were those of Rev. Proc. 2017-58; after the Tax Cuts and Jobs Act, Rev. Proc. 2018-18 (5 March 2018) superseded them.

\section{Terms and legal background}

\subsection{Terms}

\textbf{Terms used in this paper.} The paper is written for readers from law and from computer science, so we define a few terms. Where the two fields use different words for the same thing, we give both.

\begin{center}
\footnotesize
\begin{tabularx}{\linewidth}{p{0.22\linewidth}X}
\toprule
Term & Meaning here \\
\midrule
As-of-date decision & Deciding a case under the law that applied on a given date, not the law today. Lawyers speak of the law applicable in time (intertemporal law); in databases this is a point-in-time query. \\
Executable rule & A program that takes the facts of a case and returns the legal outcome (a status or an amount); here a Python function. Also called "rules as code". \\
Answer key (key) & Outcomes recorded by others, independently of the author's reading: firms' own disclosures, a tax calculator's results, insurers' decisions. Computer scientists may call this ground truth; we avoid the term because a key can itself be wrong (Section 8). \\
Reference rule & The author's hand-written executable rule: one reading of the law, scored against the key like any other rule, not the correct answer. \\
Gold labels & Labels written by the author for a component, such as which provisions a decision needs (Appendix A). Unlike the key, they are not independent. \\
Version store & Every historical version of each legal text, with the dates from which it was in force. \\
Delegation & A higher-ranking text leaving details to a lower-ranking one: an act to a decree and its annex (Korea), the Code to yearly Revenue Procedures (US), the Social Code (SGB V §136b) to the Federal Joint Committee's minimum-volume rule, Mm-R (Germany). \\
Addendum (Korean 부칙) & The closing provisions of an amending act or decree, stating when the amendment takes effect and to whom it applies (Section 2). \\
Transitional provision & A provision that keeps an earlier legal position for a stated time after an amendment. A deeming provision is one kind: the law treats a firm as an SME for a period regardless of its figures. \\
Grace period & The three years for which a firm that outgrows the SME ceilings keeps SME status (Act Art. 2(3)). \\
Application period & The year, 1 April to 31 March, for which a firm's SME status is fixed by the previous fiscal year's figures. \\
Firm-year & One firm's annual report for one fiscal year; the unit of the Korean evaluation. \\
Retrieval; retrieval condition & Choosing which provisions to give the model. The conditions in Section 5.2 differ only in this; the oracle condition gives exactly the provisions a decision needs. \\
Synthesis & An LLM writing an executable rule from the retrieved texts. \\
Property; repair & A statement every correct rule must satisfy (for example, a firm deemed an SME is an SME in that year), checked on generated inputs without the key; repair asks the model to fix a rule that violates one (Section 5.3). \\
Agreement & The share of cases in which a rule and a key give the same outcome. \\
Pre-registered criterion & The decision rule for declaring a difference between conditions, fixed in the analysis plan before LLM outputs were scored (Section 5.2). \\
Entitlement (German) & A hospital's permission, for one year, to perform and bill a procedure subject to a minimum volume (Appendix F.1). \\
\bottomrule
\end{tabularx}
\end{center}

\subsection{Legal background in detail}

\textbf{Korean legislation for readers outside Korea.} Korean legislation is layered. An act (법률) passed by the National Assembly sets the frame and delegates details to a presidential enforcement decree (시행령) and, below it, to ministerial rules (시행규칙); a decree may put figures such as ceilings into annexes (별표), which are part of the decree and amended with it. Every amending act or decree ends with addenda (부칙). Addenda are part of the amending instrument and have the same rank: an addendum to a decree binds as the decree does. They typically hold an effective-date article, application rules (적용례) that tie the new rule to facts or periods, transitional measures (경과조치) that keep an earlier legal position for a stated time, and special rules (특례). Addenda are not merged into the amended text; the consolidated text carries the addenda of every amendment as a chronological appendix, so a reader of the current version still sees provisions whose effect has expired and must determine which of them still apply. Interpretations by the Ministry of Government Legislation (법제처 법령해석) answer questions from agencies or citizens on the meaning of a provision. They are dated, published, and generally followed by the administration, but they do not bind the courts, and the Supreme Court decides the meaning of a provision with binding effect only in the case before it. For SME status, the Ministry of SMEs and Startups also issues certificates on request, but firms state their own status in their annual reports, and that statement is our key.

\textbf{Transitional provisions and constitutional doctrine.} Korean drafting places transitional provisions in the addenda of the amending instrument, as effective-date articles, application rules, transitional measures, and special rules [Ministry of Government Legislation 2017]. Under the doctrine of legitimate expectations, where a new rule reaches facts that began under the old one (non-genuine retroactivity), the Constitutional Court weighs the reliance interest against the public purpose of the change, and applies stricter review than to purely prospective rules [Constitutional Court 1995]; German constitutional doctrine draws the same line between genuine retroactivity (\emph{Rückbewirkung von Rechtsfolgen}) and non-genuine retroactivity (\emph{tatbestandliche Rückanknüpfung}) [BVerfG 1986]. Where the legislature has drawn the boundary of a transitional provision, the Court treats it as a legislative choice and expectations beyond it as mere expectations, not protected reliance [Constitutional Court 2019]. How far a general transitional clause reaches is itself litigated and debated, for example whether its "previous provisions" can be read literally once the surrounding regime has changed [Kim 2020].

\textbf{The temporal life of an addendum.} The Supreme Court holds that a whole amendment in principle extinguishes the old addenda, but that an old transitional provision continues to apply where its history and purpose, and the structure of the new act, show that the legislature meant it to, and where continuing it imposes no unforeseeable burden on those it addresses; applying that test, it held that an addendum to a tax act remained in force [Supreme Court 2008; Supreme Court 2013]. The Constitutional Court held that reading unconstitutional, because a lapsed addendum cannot be revived by interpretation as a basis for taxation [Constitutional Court 2012], and the Supreme Court in turn held that such a decision on an interpretation does not bind the courts [Supreme Court 2013]; commentators read the conflict as one over the limits of judicial law-making [Lim 2012; Kang \& Kim 2014]. The drafting standards ask drafters of a whole amendment to check which earlier addenda must be carried over.

\section{Execution and audit details}

\textbf{Models and clients (Section 5.4).} The main runs use Gemini 3.8 Flash (identifier gemini-3.8-flash-high) through the Antigravity command-line client 1.2.14, between 8 and 9 November 2026. The client is agentic: it can read files, run commands, search the web, and start sub-agents, so isolation has to be enforced and verified. Claude Opus 5.5 answered the same 343 prompts on 9 November 2026 through Claude Code subagents (scripts/opus\_collect.py; audit in Table C1). Two further models were run the same day: GPT-OSS 120B, an open-weight model, through the same Antigravity runner and audit as Gemini, and Claude Haiku 5.5 through subagents as for Opus 5.5. For Haiku, 108 subagents first responded to a chat message that had reached the running subagents instead of doing the task; they were rerun with the same instructions, and all 451 Haiku transcripts stayed inside their own folders (data/haiku55\_audit\_note.md). The 16 graded responses whose first code block did not load were regenerated as the analysis plan prescribes (the 17th, US 2023, lies outside the graded years; Table C1).

\textbf{Procedure.} \emph{Main runs (Gemini 3.8 Flash).} Each prompt runs in a new empty directory outside the project. In the first isolated runs the prompt was the only file there; later it was passed on standard input, and the run-level settings denied file reads, writes, and commands. Web search cannot be disabled through the client, so every prompt tells the model not to use tools, and every conversation log is audited for project paths, answer-key file names, search queries, and external URLs. A conversation with a project path, an answer-key file name, or web use (a search query, or more than three distinct external URLs) is discarded and regenerated. Each run is limited to 10 minutes and retried up to three times if the response has no valid function, that is, if its first code block does not load and define \texttt{judge}; this is the analysis plan's rule, applied after the audit (the scorers had earlier dropped such responses silently). \emph{Second model (Claude Opus 5.5).} One folder per prompt outside the project; the subagent is instructed to read only that file and to write its answer there, and every transcript is audited for tool calls (Section 5.4).

\begin{table}[t]
\caption{Audit counts (data/agy\_audit\_record.md, data/agy\_runs.csv, data/opus55\_audit.csv).}
\footnotesize
\begin{tabularx}{\linewidth}{p{0.22\linewidth}XXX}
\toprule
Run & What was checked & Result & Action \\
\midrule
First round, inside the project directory & 78 conversation logs & 33 name evaluation data files (e.g. \texttt{cases.csv}) that no prompt mentions & whole round withdrawn \\
First isolated runs & 46 + 26 conversations (the two windows overlap by minutes) & no project path, no answer-key file name & kept \\
Rerun of 206 runs & 207 conversations in its window & the 206 runs inside isolated directories, none naming an answer-key file; one non-synthesis conversation from the project directory & kept \\
All Gemini runs & web use & 2 conversations ran a web search (one for an Annex 1 ceiling) & responses discarded and regenerated \\
All Gemini runs & attempts logged before the audit & 525 of 814 ended in a client error (mostly the usage quota, where observed) & retried \\
All Gemini runs & first code block loads & 17 of 303 synthesis responses did not (Korea 14, US 2, Germany 1) & 16 regenerated, 14 of which now load; the 17th is US 2023, outside the graded years \\
Last rerun before the audit & 33 conversations & no project path, answer-key file name, search query, or external URL & kept \\
Second model (Claude Opus 5.5) & 343 subagent transcripts & every run only read its prompt and wrote or edited its own answer (two through a shell heredoc to their own file) & 5 runs blocked by a transient permission-check failure were rerun \\
Third and fourth models (GPT-OSS 120B; Claude Haiku 5.5) & 343 Antigravity conversations; 451 subagent transcripts & GPT-OSS: all valid, 7 client errors retried; Haiku: every path inside the run's own folder; 108 runs first answered a chat message instead of the task & GPT-OSS: retried; Haiku: 108 rerun \\
\bottomrule
\end{tabularx}
\end{table}

\textbf{The withdrawn round and withdrawn material.} On the 90 firm-years answered without code in both rounds, the no-law condition stated the correct revenue ceiling in 100\% of cases in the withdrawn round and in 60\% after isolation, and agreed with the reference rule in 94\% against 86\%; the conditions given our bundle (70 cases) or a date-filtered corpus (90 cases) gave identical results in both rounds. The leakage inflated performance exactly where the model lacked information. Agents tampering with or leaking into their own evaluation, and contamination inflating legal reasoning scores, have been measured elsewhere [Atinafu \& Cohen 2026; Kordjamshidi et al. 2026]; this is an instance of the problem in rule synthesis. Those cases belong to a direct-answer probe (40 of the 343 prompts) that was run and then dropped from the results because of a design flaw: its input omitted pre-2015 status, so the 33 cases drawn for the 2014 addendum were unusable.

\textbf{Label extraction (Section 4.1).} The extractor matches with whitespace removed, reads negations such as "해당하지 아니합니다", classifies the clause that cites the Framework Act when a sentence also cites a tax statute, and treats a statement that the firm is in its grace period as SME status; "not applicable" (해당사항 없음) in the SME field is read as non-SME. The disclosed status is the status "as of the filing date of this report" (the wording of the disclosures). The 82 disclosures that cite the SME definition of a tax statute cite the Restriction of Special Taxation Act or the Corporate Tax Act, and 15 of them do not cite the Framework Act; they are marked in the released labels (column law).

\textbf{Financial facts (Section 4.1).} The first-filed rule holds only in part: figures for FY2013–2014 come from the comparative columns of FY2015 reports, and some later years are available only from a later report. Firms with a short listing history are averaged over the years available. Total assets are treated as missing when implausible for a listed firm: below KRW 1 billion or above KRW 20 trillion, or off the firm's other years by a factor of 300 or more (scripts/fin\_sanity.py). Annex 2 of the decree adds to a parent's revenue the revenue of subsidiaries in which it holds 50\% or more, and a proportional share of the revenue of those held below that; consolidated statements approximate this aggregation for majority-held subsidiaries. Using separate-entity figures lowers agreement with the disclosures from 89.1\% to 86.3\% overall (on 11,079 firm-years with separate figures) and from 76.8\% to 68.8\% within ±20\% of the ceiling. Each firm carries one industry code, its current one, for all years. By industry, the 10,901 firm-years are distributed as follows: manufacturing 66\%, information and communication 15\%, wholesale and retail 7\%, professional and scientific services 5\%, construction 2\%, and other industries 5\% (data/sme/kr\_numbers.md).

\textbf{US key (Appendix F.1).} The hand-coded 2018 rule's match with TAXSIM confirms that the key computes what the task specifies. A function can take the AMT exemption amounts from the Revenue Procedure, but the structure of the AMT computation must come from the model's own knowledge; we keep the store as it was when the runs started. Our task specification asks for income tax before credits including AMT, but TAXSIM reports AMT on line 27 and leaves it out of line 28; AMT applies to 9–20 returns per year in 2013–2017 and to none afterwards. The statute reduces personal exemptions by two percentage points for each \$2,500 or fraction thereof, whereas TAXSIM phases them out continuously. For 2023, for a single filer aged 66 with wages of \$589,370 and interest of \$3,630, Rev. Proc. 2022-38 gives a standard deduction of \$15,700 and tax of \$173,949.50. TAXSIM35 (build of 5 March 2024) instead returns a deduction of \$15,600, which is the 2023 basic deduction (\$13,850) plus the 2022 additional deduction for age (\$1,750), and tax of \$176,593.00, which is the 2022 rate schedule applied to taxable income of \$577,400. Documentation of earlier TAXSIM versions states that indexed parameters beyond the coded years are extrapolated. Disagreement across synthesis conditions is a practical way to find the coverage limits of an independent key.

\textbf{Retriever and addenda reader (Section 5.1).} The retriever excludes addenda whose stated end dates have all passed. A Korean addendum bundles articles of different kinds: an effective-date article, application rules, transitional measures, special rules, and amendments to other acts. The reader splits each addendum into articles and reads, from the article title and text, (i) its kind; (ii) the provisions it mentions, normalized to article keys; (iii) its start (promulgation date, an explicit effective date, or "six months after promulgation"); and (iv) its end (an explicit "until" date, "three months after the end of the fiscal year containing the effective date", or "three months from the start of the first fiscal year"). An article is included for decision date \emph{d} if it is substantive (application rule, transitional measure, special rule, validity period), alive on \emph{d}, and mentions a provision already in the bundle.

\textbf{Synthesis inputs (Section 5.2; Appendix F.2).} The padded Korean list reaches 89–90\% of our bundle's length in 2016–2018, where too few unrelated provisions exist. The German Mm-R inputs omit the procedure-code lists. There is no delegation walk for Germany because a single rule document carries the thresholds, transitional clauses, and the review standard.

Results are reported with the regenerated functions and, as sensitivities, with the original responses left out or counted as wrong on every case (Appendix D.2).

\section{Detailed results}

\subsection{Disagreements between the reference rule and the disclosures}

\textbf{What remains.} We examined the remaining 11\% (1,183 firm-years) through a stratified sample of 60 disagreements, 15 for each combination of direction and label format, drawn before the label corrections and classified by the author from the report text and from a rerun of the rule with separate-entity figures (scripts/disagree\_sample.py, scripts/disagree\_classify.py; data/sme/disagree\_summary.md). In the sample, six cases were label extraction errors and one was a unit error in the financial facts; the corrected labels and the financial filter resolve all seven, and one further case now agrees. The 52 cases that still disagree are weighted to represent the 89\% of current disagreements that were already disagreements in the same stratum before the corrections:

\begin{itemize}
\item about 32\% are firms below the ceiling that disclose non-SME status, which points to the independence requirements the rule omits (Decree Art. 3(1)(ii), Annex 2); one belongs to an agricultural cooperative group, another is a holding company;
\item 27\% are firms that the rule places in a grace period while they disclose non-SME status, which fits a grace period used before our data begin and the once-only rule of Decree Art. 9(iii) (one firm states that its grace period ended in 2005);
\item 15\% agree with the disclosure once separate-entity figures replace consolidated ones, so the revenue basis chosen to approximate Annex 2 decides them;
\item 5\% lie within 10\% of the ceiling;
\item 7\% turn on whether the 2014 addendum applied, including the firm's pre-2015 status;
\item 15\% disclose SME status with figures above the ceiling on both bases and no grace period in our data: reporting errors, a different industry classification, or a grace period we cannot see.
\end{itemize}

The other 11\% of current disagreements were not in the sampling frame; 100 of these 128 are firms that the rule places below the ceiling and that answer "not applicable" in the SME field, which is consistent with the independence group. Missing facts thus account for most of the gap, and the revenue basis and the 2014 addendum for a smaller part. The sample no longer shows label errors, but it was not drawn to estimate them; the residual error of the key is what the 200-label check measures. The two largest groups concern the provisions on which the Ministry of Government Legislation has issued interpretations (independence and related firms; the grace period; Section 2). The classification is the author's, made without ownership data or certificates.

\subsection{Synthesis results in detail}

Results from all 303 synthesis outputs, after the regeneration of 9 November (Appendix C). Of the 16 regenerated responses, 14 load; the two Korean no-law responses for FY2020 still failed after three attempts and are left out, which leaves that cell with one function. Synthesized functions are scored as fixed in the analysis plan: mean and range over three functions per cell, and a difference counts only if the mean difference exceeds the larger within-cell range and at least 7 of 9 pairwise comparisons point the same way. A response is valid if its first code block loads and defines \texttt{judge}; a function that runs but fails on a case counts as wrong on that case. Every result below was also computed with the original responses left out or counted as wrong on every case, with the two remaining invalid responses counted as wrong, with the last instead of the first code block, and, for Korea, only on cases whose filing date lies in the window stated in the prompt, under reading (a), and on the labels before the audit (scripts/score\_all.py; files \texttt{data/*/synth/score\_\_*.md} and \texttt{criterion\_\_*.md}). Where a setting changes a verdict we say so.

\begin{table}[t]
\caption{Pre-registered primary comparison, our bundle against the date-correct fixed list, per year and model ("yes" = the criterion of Section 5.2 is met in favour of our bundle). Last column: the RQ2 verdict across models, for which a difference needs at least three of four models (analysis plan, change log of 9 November 2026). Subset A: firms covered by the 2014 addendum, FY2015–2018; subset B: all firms, FY2017–2019 (data/four\_models.md).}
\footnotesize
\begin{tabularx}{\linewidth}{p{0.22\linewidth}XXXXXXX}
\toprule
Subset & FY & Key & Gemini 3.8 Flash & Claude Opus 5.5 & GPT-OSS 120B & Claude Haiku 5.5 & RQ2 verdict across models \\
\midrule
A & 2015 & reference rule & no & yes & no & yes & model-dependent (2 of 4) \\
A & 2016 & reference rule & no & no & no & yes & model-dependent (1 of 4) \\
A & 2017 & reference rule & no & yes & no & yes & model-dependent (2 of 4) \\
A & 2018 & reference rule & no & no & no & fixed list ahead & none (0 of 4) \\
A & 2015 & disclosures & no & yes & no & yes & model-dependent (2 of 4) \\
A & 2016 & disclosures & no & no & no & yes & model-dependent (1 of 4) \\
A & 2017 & disclosures & no & yes & no & no & model-dependent (1 of 4) \\
A & 2018 & disclosures & no & no & no & fixed list ahead & none (0 of 4) \\
B & 2017 & reference rule & no & yes & no & no & model-dependent (1 of 4) \\
B & 2018 & reference rule & no & no & no & no & none (0 of 4) \\
B & 2019 & reference rule & no & no & no & no & none (0 of 4) \\
B & 2017 & disclosures & no & yes & no & no & model-dependent (1 of 4) \\
B & 2018 & disclosures & no & no & no & no & none (0 of 4) \\
B & 2019 & disclosures & no & no & no & no & none (0 of 4) \\
\bottomrule
\end{tabularx}
\end{table}

No sensitivity setting turns a Gemini verdict into a difference, with one exception: on the labels before the audit, our functions pass the criterion in FY2015 of subset A against both keys (1.000 against 0.974 and 0.925 against 0.896). With the original responses left out, FY2015 and FY2018 of subset A have fewer than three Gemini functions in one condition and cannot be judged.

In Table D1, the eight rows in which the criterion is met for one or two models all concern the 2014 addendum firms of subset A or in FY2017 of subset B.

\textbf{Korea: pooled shrinkage and the null ratio (Section 6.3; exploratory).} On the firm-years with both a reference decision and a label, our bundle's lead against the disclosures is 0.28 (padded list), 0.29 (fixed list), 0.50 (no law), and 0.60 (current law) of its lead against the reference rule (data/robust\_null.md; Table 7). The null ratio 2p − 1 = 0.78 was defined after the Gemini and Opus results had been seen and fixed in advance only for the other two models. For Gemini every ratio lies below 0.78, including the upper end of its 95\% interval; for Opus two of four do and a third sits at the boundary (data/robust\_null\_\_opus55.md; next paragraph). Without stratification, splitting by whether the reference rule is right gives the following. On the 89\% of firm-years where the reference rule agrees with the disclosures, our bundle leads the fixed list by 0.015 (0.009–0.022) and no law by 0.041 (0.032–0.051); on the 11\% where it disagrees, by 0.068 (0.042–0.096) and 0.112 (0.080–0.148) against the reference rule, leads that on these firm-years are losses against the key. Those firm-years carry a quarter to a third of the lead against the reference rule. Because harder firm-years are over-represented where the reference rule disagrees, Section 6.3 reports the comparison within strata of difficulty (data/robust\_strat.md) instead of this unstratified split. Part of the shrinkage also reflects a ceiling set by the reference rule's own agreement with the disclosures, and part reflects that the primary subset separates the legal routes poorly (the grace route, Section 5.2; next paragraph: subset A, a post-hoc FY2019–2020 subset, and reading (a)).

\textbf{Korea: subsets, routes, and reading (a).} On subset A, where cells are small and ranges wide, the pooled gaps shrink less than over all firm-years (fixed list 0.044 and 0.040; no law 0.016 under both keys). For Opus, of the four ratios against the null ratio of 0.78, two lie below it, a third sits at the boundary (upper end 0.782 against 0.783), and the fourth, against no law, rests on a lead of 0.008 too small for a ratio to mean much (data/robust\_null\_\_opus55.md). In a post-hoc analysis of FY2019–2020, where the legal routes diverge, the conditions do not differ against the disclosures either (0.64 against 0.63; 0.81 against 0.80), close to the 0.70 that the reference rule itself reaches on addendum firms in FY2019. Under reading (a) of the addendum the ranking against the reference rule is unchanged (ours and the oracle 0.979, fixed list 0.960, current law 0.951, no law 0.936).

\textbf{Four models (Section 6.3).} Under rules fixed before the third and fourth models were scored (analysis plan, change log of 9 November 2026), all 343 prompts were run again with Claude Opus 5.5, GPT-OSS 120B, and Claude Haiku 5.5 (Section 5.4; data/four\_models.md); Table D1 gives the pre-registered verdicts, and the rest of this paragraph is exploratory. The two Claude models do not reach the addendum's outcome by the route Gemini takes (the grace route, Section 5.2): given the fixed list, which lacks the addendum, their functions agree with the reference rule on 0.585 (Opus) and 0.440 (Haiku) of subset A, against 0.851 for Gemini (see the next paragraph for a reading of the functions and an ablation of the pre-2015 field). Without that route, the provision the bundle adds decides the outcome, and the lead survives the independent key (Opus, subset A, FY2015: 0.935 against 0.348 against the disclosures). Pooled over years, the lead of our bundle shrinks against the key for three models; for Haiku it shrinks beyond the null ratio in one of four comparisons, and its larger leads (0.092 over the fixed list against the reference rule, 0.071 against the disclosures) mostly survive. GPT-OSS 120B is the exception: its functions are far from both keys (our bundle 0.651 against the reference rule and 0.638 against the disclosures; no law 0.748 and 0.766), and its differences do not shrink (against no law it trails: −0.103 against the reference rule, −0.129 against the disclosures); one possible but untested reason is that its errors are not concentrated where the author's reading is wrong. Lacking the amounts in memory, it gains from retrieval in both comparison cases, as does Haiku 5.5 in the US (Appendix F.4). Where a model already knows the law, the gains are small and partly reflect better reproduction of the author's reading.

\textbf{Korea: how the models reach the addendum's outcome.} Reading the functions shows the route (data/pre2015\_usage.md): given the fixed list, 27 of 36 Gemini functions (36 = 12 decision years × 3 samples; Table 7 scores the 27 of the labelled years) read the input field on pre-2015 status, which only the addendum makes relevant, against 5 of 36 for Opus; given current law, all three Gemini functions read it and no Opus function does. Removing the field from the prompts confirms the route for Gemini (data/synth/score\_\_nopre\_gemini.md): on subset A its fixed-list functions fall from 0.851 to 0.478 against the reference rule (0.777 to 0.506 against the disclosures) and current law from 0.875 to 0.585 (0.805 to 0.596), so their agreement with the addendum's outcome came through the field. Our bundle falls as well, as it must, since the addendum requires pre-2015 status (0.895 to 0.489; 0.817 to 0.504); without the field the two conditions end level. For GPT-OSS 120B the same ablation lowers our bundle (0.855 to 0.762 against the reference rule, 0.795 to 0.653 against the disclosures) and current law (0.609 to 0.531; 0.602 to 0.536), while the fixed list does not fall (0.773 to 0.848, within a range of 0.36–0.95) (data/synth/score\_\_nopre\_gptoss.md).

\textbf{Korea: the controls.} These comparisons were not pre-registered as primary and are descriptive. Over all firm-years our retrieval reaches the oracle (0.989 against the reference rule and 0.890 against the disclosures in both), so it recovers what the decisions need. On subset A the four conditions that differ only in retrieval rank in the same order against both keys: oracle, ours, fixed list, padded list. Padding the fixed list with unrelated provisions to the size of our bundle lowers agreement (0.776 against 0.851 on subset A), consistent with the selection of provisions mattering more than the amount of text; this bears on RQ2. Ranges within cells are wide (the padded list ranges from 0.27 to 1.00 on subset A), which is why the pre-registered criterion finds no difference. Functions written without law score lowest over all firm-years against the reference rule (0.939) and are level with current law against the disclosures (0.866 and 0.863), but the gap to ours (0.050) is smaller than the range within the no-law cell, and on subset A they score above the fixed list: memory of the pre-2015 regime and of the grace period carries the model as far as an incomplete bundle does. Current law reaches 0.875 on subset A through the grace route described above; earlier drafts reported 0.997 on a subset that left out FY2018, the year in which that route diverges from reading (b).

\textbf{Code-block extraction.} Taking the last instead of the first code block is a post-hoc sensitivity analysis (Section 6.4). With the original responses, 11 of the 14 Korean responses whose first block did not load then load and, in subset A, our functions pass the criterion in FY2015, while the German function that speculated about the grader fails to load. After the regeneration the last-block setting changes no verdict (data/synth/criterion\_\_lastblock.md).

\subsection{What the disclosure text says on the 2014 addendum}

Beyond the labels, the disclosure text bears on the 2014 addendum. In the evidence sentences of the 189 differing firm-years, six firms state when their status ended, all in a way consistent with reading (b): three say that their SME status or grace period ended on 31 March 2018 (the end of the deeming), and three that they ceased to be SMEs, or became mid-sized enterprises, on 1 April 2018. No firm states reading (a). We then searched the report texts of the firms whose label differs from one reading in all three years: 9 firms that disclosed SME status in all three (fitting reading (a)) and 31 that disclosed non-SME status in all three (fitting (b)) (scripts/buchik\_source\_check.py, data/sme/buchik\_source\_check.md). These are fewer than the 15 and 32 firms of Appendix D.4, which count every FY2018–2020 disclosure of the 78 firms, whereas here the readings must also differ in each of the three years. None of the 9 firms on the (a) side mentions an SME grace period or the 2014 addendum in its SME statement for FY2018–2020, and a search of the full reports for sentences that link "grace" or "addendum" to SME status finds none either (generic keyword counts do hit, for example tax deferrals or other addenda); 2 of their 27 firm-years mention an SME confirmation certificate. Their average revenue is, at the median, 1.6 times the ceiling. Eight of the 9 repeat word for word, through FY2020, the SME sentence they wrote for FY2017, while the deeming still applied; 26 of their 27 firm-years repeat the previous year's sentence, against 53 of 93 on the (b) side, where 12 firm-years mention the end of a grace period and six firms call themselves mid-sized enterprises. The (a) side thus shows no sign of a deliberate reading and a clear sign of reporting inertia, so the evidence favours reading (b); we keep both readings in the analysis because no official interpretation settles the question. Firms whose disclosures changed within FY2018–2020 or lack a label in one year were not checked.

\subsection{Robustness of the RQ1 comparisons}

\textbf{Grace-start convention by period and by firm.} Section 6.2 chose the earlier grace start after comparing both conventions with the disclosures over all years, on firms outside the 2014 addendum. Split by period afterwards, the earlier start wins in both FY2015–2019 (56 to 33 firm-years) and FY2020–2023 (224 to 57; agreement 89.4\% against 86.3\%). This split is not a held-out test, because the choice was made with both periods already in view; the held-out check is the split by firm halves in Section 6.2. Counted by firm, the earlier start also wins in both periods (FY2015–2019: 52 firms to 29, exact sign test p = 0.014; FY2020–2023: 196 to 43; data/robust\_ci.md).

\textbf{Transitions and status-fixing (Section 6.1).} The 27 firms that failed the size test in FY2015–2017 yet disclosed SME status throughout and became non-SMEs in FY2018 account for 27 of the 40 SME-to-non-SME transitions that year. Status-fixing works the same way: none of 44 firms whose revenue lies between the old and the raised 2025 ceilings switched to SME status in the FY2025 report (an observation outside the key of Table 4, which ends with FY2023), because Decree Art. 3-3 had fixed their status under the old annex for the period that began on 1 April 2025.

\textbf{Sources searched on the 2014 addendum.} The official interpretations we read on the grace period and the 2014 amendment (14-0128, 15-0157, 21-0131; Section 2) do not address grace after the deeming ends, and none of the twelve interpretations the Ministry of Government Legislation lists under the Framework Act addresses the question. The Ministry of SMEs and Startups' guide to the SME scope (2022 edition) reprints Addenda Art. 2 but, in its questions on the grace period, explains only Art. 3; its 2015 edition, listed on the ministry's site, could not be retrieved [Ministry of SMEs and Startups 2022]. A web search found no published reply of that ministry on the question (its inquiry board cannot be searched from outside). A search of the court and Constitutional Court decisions in the National Law Information Center (9 November 2026) for the provision found no decision on Paragraph 2; the decisions it returns concern other statutes' addenda.

\textbf{Firm-level patterns on the 2014 addendum.} Of the 78 firms on which the two readings differ, 32 disclosed non-SME status in all of FY2018–2020, as (b) implies, and 15 disclosed SME status in all three, as (a) implies; 21 switched from SME to non-SME within the period, 4 show another pattern, and 6 lack a disclosure in some year.

\textbf{Pre-2015 status checked against FY2012–2014 figures.} The pre-2015 status is taken from FY2014 disclosures. Checking it against the old upper limits (assets of KRW 500 billion, equity of KRW 100 billion, three-year average revenue of KRW 150 billion) overrules 35 of those disclosures (Table 6, last row). The readings then differ on 121 firm-years, with the disclosures siding with (b) in 85 and with (a) in 36 (by firm, 36 to 18, p = 0.02). The old limits applied to a firm's own accounts, while our figures are consolidated where available, so this check may overrule too many, and the employee limit cannot be checked.

\textbf{Grace counted one report later.} Counting the grace period one report later (FY2019–2021) gives 88.2\% overall and a split of 45 (a) to 153 (b) on 198 firm-years.

\textbf{Coverage and "not applicable" labels.} The split does not rest on how coverage is defined or on the "not applicable" labels. Restricted to firms that failed the new size test in FY2015, when the new criteria took effect, it is 52 to 114 (reading (a) first); without the "not applicable" labels, 59 to 113 by the extractor's marking and 59 to 111 by the author's hand check, which finds 19 such disclosures rather than 17.

\textbf{Before the label audit.} Before the audit's label corrections the split was close (104 to 113); the same corrections, which nearly all concern firms whose sentences denied SME status in forms the earlier extractor missed (for the repair property mostly from FY2018 on), also reversed the result of the repair property (Appendix D.5).

\textbf{Firm groups.} The firm groups in this paper are distinct: the 63 firms to which the addendum applied in all of FY2015–2017 (and disclosed in those years), the 98 firms that failed the size test yet disclosed SME status throughout FY2015–2017 (Section 6.1), and the 78 firms on which the two readings of the 2014 addendum differ (Section 6.2).

\subsection{The verify-and-repair loop}

In the first isolated runs, the property for the 2014 addendum (Table 5) flagged the rule synthesized from current law that had reproduced the addendum's outcome through the grace period: it treats covered firms as SMEs in the 2020 period, which reading (b) excludes. The test found this without consulting the disclosures, and one repair round removed the violation (10 of 20 property checks failing before, 0 after, with no new violations). The properties flagged five synthesized functions in all: the three functions built from current law and one oracle function for FY2020 (all on the 2014 addendum property), and one German function built with our bundle, the one whose first code block was a fragment (Section 6.4); no US function violated a property. Scored once after the loop, the first current-law function's agreement with the reference rule rose from 93.8\% to 94.6\% overall and from 89.7\% to 94.0\% on firm-years from FY2018 onwards of firms that were SMEs before 2015 (3,111 firm-years); its agreement with the disclosures rose from 86.5\% to 86.9\% overall and from 88.0\% to 90.0\% on that subset (3,030 firm-years with a label). Where the repair changed the decision, the disclosures sided with the repaired function in 215 cases and with the unrepaired one in 170. Before the label corrections of Section 8 the direction was reversed: on the earlier labels the disclosures sided with the unrepaired function (212 against 192) and its agreement with them did not rise. After the corrections, the other two current-law functions moved the same way as the first (reference rule 0.949 to 0.952 each; disclosures 0.861 to 0.863 each; changed cases 63 against 46 and 61 against 42), and so did the oracle function for FY2020 (disclosures 0.859 to 0.887; changed cases 50 against 16); the repaired German function reaches balanced accuracy 0.863 in 2021, where the original had returned no value on 88\% of cases. The property encodes reading (b) of a contested provision, and repair imposes that reading; whether this helps against the key depends on whether the key follows the same reading and on the quality of the key itself, so we make no claim for the loop and report its results with and without this property (scripts/repair\_vs\_gold.py; data/metamorphic\_*\_repair1.md).

\section{Datasheet}

Following Gebru et al. (2021), Table E1 answers the standard datasheet questions for the released test bed.

\begin{table}[t]
\caption{Datasheet for LexTemporis.}
\footnotesize
\begin{tabularx}{\linewidth}{p{0.22\linewidth}X}
\toprule
Question & Answer \\
\midrule
Purpose & To grade executable rules for as-of-date legal status decisions against keys independent of the reading that produced them. Not built to decide any firm's actual SME status. \\
Creator and funding & Withheld for double-blind review; no specific funding. \\
Instances & Korea: 10,901 firm-years (1,677 listed firms, FY2015–2023), each with a disclosed SME label, revenue and assets, industry code, pre-2015 status, and the filing identifier. US: 2,000 synthetic returns (2013–2022) with TAXSIM results. Germany: 7,845 site–procedure–year decisions (entitlement years 2021–2024) with case counts, exceptions, and the insurers' result. Version stores: every version of the governing Korean act, decree, annexes, and addenda; US Code editions and Revenue Procedures; 24 Mm-R versions. \\
Labels and their source & Korean labels are the firms' own statements in annual reports, extracted by scripts/extract\_sme\_labels\_v3.py; 526 rest on "not applicable" table cells and 82 cite a tax statute (both flagged). US labels are TAXSIM35 outputs. German labels are the insurers' decisions as published by IQTIG. \\
Known errors and gaps & An audit found about 2\% of first-extractor labels wrong (corrected in version 3); the author's hand check of 200 stratified labels found 2 errors in 183 decidable ones, and confirmed every decidable label of the 189 firm-years on which the readings of the 2014 addendum differ; a second annotator, an undergraduate not involved in the project and blind to labels and the author's decisions, agrees with the author on 352 of 353 disclosures both decided (κ = 0.99) and differs mainly in coding "not applicable" as non-SME against the instructions (28 cases). Financial figures failing a sanity filter (271 rows) are treated as missing. FY2015 revenue averages two years, not three. The Korean reference rule omits the ownership-based independence tests. \\
Personal data & None. Firms are legal persons identified in public filings; the US returns are synthetic; German data are at hospital-site level. \\
Collection & Korean filings via the DART Open API and legal texts via the National Law Information Center API (2026); TAXSIM35 run locally (version of 5 March 2024); IQTIG result files for report years 2019–2022; Mm-R versions from the G-BA. \\
Preprocessing & Label extraction (version 3), financial sanity filter, Annex 1 parsed into (industry range, ceiling) rows per version, addenda split into articles. All scripts are released. \\
Intended and unsuitable uses & Intended: evaluating rule synthesis, retrieval, and encodings against independent keys; studying how the date enters legal decisions. Unsuitable: certifying a firm's SME status, tax advice, or hospital entitlement decisions. \\
Distribution and licence & Released with the paper at https://github.com/nepersoned/lextemporis (scripts, prompts, responses of all four models, derived data, version stores). Code under MIT; the author's own derived data (labels, reference decisions, prompts, model responses, scores) under CC BY 4.0. Third-party material keeps its own status: Korean statutes are not protected by copyright (Copyright Act Art. 7), US federal texts are in the public domain, and the short excerpts of DART business reports remain the companies' and are included only for verifying the labels; IQTIG case tables will be released only after redistribution terms are confirmed (see Data and code availability). \\
Maintenance & Maintained by the author; versioned releases with a changelog (paper/CHANGELOG.md). New fiscal years and law versions can be added with the collection scripts. \\
\bottomrule
\end{tabularx}
\end{table}

\section{Comparison cases: US income tax and German hospital minimum volumes}

The two comparison cases meet the selection criteria of Section 1 each with a different kind of key and a different legal layer carrying the date. This appendix holds their data, conditions, and results; the main text refers to them where they bear on the Korean case.

\subsection{Data and keys}

\textbf{US income tax.} 200 synthetic returns per tax year 2013–2022 (single or joint filers; wages, interest, qualified dividends, long-term gains; ages around 65 to trigger the additional standard deduction); 2023 was generated but dropped (below). The key is income tax before credits from NBER TAXSIM35 (local executable, version of 5 March 2024), its maintainers' interpretation of the law, independent of ours; a hand-coded 2018 rule matches it within \$1 on all 200 returns of that year. The version store holds IRC §§1, 63, 151 (GovInfo annual editions 2012–2023) and the 13 Revenue Procedures setting the inflation-adjusted amounts for 2013–2024, but not §55, so the structure of the alternative minimum tax (AMT), which the task includes, must come from the model's own knowledge; we report this gap rather than change the store.

\textbf{Checking the US key.} Three discrepancies surfaced only when synthesized rules were compared with TAXSIM across tax years, and were resolved before analysis (details in Appendix C). TAXSIM reports AMT on a separate line (27) and leaves it out of income tax before credits (line 28), so we grade against their sum; it phases out personal exemptions continuously rather than in the statute's steps, a difference of \$2–\$52 within the 1\% tolerance we report alongside exact agreement; and for 2023, outside its verified coverage, it mixes 2023 and 2022 parameters, so we grade 2,000 returns for 2013–2022.

\textbf{German hospital minimum volumes.} IQTIG publishes, for report years 2019–2022, every hospital site and procedure with the case volumes used in the prognosis, the insurers' review result, exceptions (new or resumed service, special permission by the state), and the resulting entitlement two years ahead; entitlement is reproduced from the review result and the exceptions in 99.4\% of rows. The version store holds 24 versions of the Federal Joint Committee's minimum-volume rule (Mm-R) from July 2019 to August 2026.

\subsection{Conditions and properties}

The conditions follow Section 5.2. US: no law; the year's statute edition; the statute plus that year's Revenue Procedure; one function from the current statute and latest Revenue Procedure for all years. German: no law; the Mm-R version in force on 31 July of the year before the entitlement year (the prognosis deadline); one function from the current version for all years. With no delegation walk, the German comparison tests version selection only (Appendix C). Each condition-year cell is sampled three times, as in Korea.

The analysis plan fixed the primary comparisons for these cases (US: ours versus statute only; Germany: ours versus current version), the German primary subset (years in which a threshold changed), and the criterion of Section 5.2. For the US, an advantage of our bundle is claimed only if the no-law condition falls clearly below it; otherwise we report that the model already knows the year's amounts.

\begin{table}[t]
\caption{Properties used by the verify-and-repair loop in the comparison cases (Section 5.3).}
\footnotesize
\begin{tabularx}{\linewidth}{p{0.22\linewidth}XX}
\toprule
Case & Property & Source \\
\midrule
US & Zero tax below any standard deduction & IRC §63 \\
US & Tax non-decreasing in wages; marginal increase below 50\% & IRC §1 \\
US & Long-term gains never taxed above wages & IRC §1(h) \\
US & Joint return with a non-earning spouse never above single & IRC §§1, 63 \\
Germany & Entitlement non-decreasing in case counts & Mm-R §4(1) \\
Germany & Exceptions grant entitlement; counts above every threshold grant entitlement & Mm-R §§4, 6 \\
\bottomrule
\end{tabularx}
\end{table}

\subsection{Where the date enters}

\textbf{US: the amounts.} Under inflation indexing the amounts change every year (Table F2).

\begin{table}[t]
\caption{Same taxpayers, law of another year (TAXSIM, 2,000 returns, tax years 2013–2022).}
\footnotesize
\begin{tabularx}{\linewidth}{p{0.22\linewidth}XXXXX}
\toprule
Law applied & Returns & Correct within \$1 & Tax differs by ≥1\% & Median error & Median error (\%) \\
\midrule
Previous year (Y−1) & 2,000 & 8.2\% & 49.7\% & \$151 & 1.1\% \\
Latest (2022) & 1,800 & 8.2\% & 90.5\% & \$1,062 & 14.1\% \\
Last pre-TCJA year (2017) & 1,800 & 6.9\% & 75.4\% & \$540 & 11.8\% \\
First TCJA year (2018) & 1,800 & 8.4\% & 86.8\% & \$812 & 11.5\% \\
\bottomrule
\end{tabularx}
\end{table}

With the binary criterion used for Korea (the tax changes by 1\% or more), the previous year's law changes 50\% of returns and the latest law 91\%, against 3.5\% of Korean decisions changed by a later annex; the latest delegated document before a date governs the wrong year (Appendix A).

\textbf{Germany: transitional clauses and an evaluative review standard.} Each Mm-R annex entry states a final threshold at its head and the current years' threshold in a transitional clause at its end (breast cancer surgery: final 100, "no minimum in 2022 and 2023, 50 in 2024"); the IQTIG files print the transitional values.

\begin{table}[t]
\caption{Threshold rule versus the insurers' decisions (7,845 site–procedure–years). A site is predicted entitled if an exception applies or a case count reaches the year's threshold.}
\footnotesize
\begin{tabularx}{\linewidth}{p{0.22\linewidth}XX}
\toprule
Count used & Agreement & Recall of denials (648) \\
\midrule
Report year & 0.864 & 0.863 \\
July–June & 0.864 & 0.920 \\
Either & 0.898 & 0.852 \\
Both & 0.830 & 0.931 \\
Predict "entitled" for all & 0.917 & 0.000 \\
\bottomrule
\end{tabularx}
\end{table}

Because 92\% of decisions grant entitlement, we use balanced accuracy (the mean recall of grants and denials). The threshold rule reaches 0.880 averaged over years (range 0.856–0.914; denial recall 0.853), against 0.500 for a constant prediction, yet its accuracy is lower than the constant's. The gap lies among sites below the threshold without an exception (Table F4). Under the Mm-R a hospital's prognosis confers entitlement unless the insurers refute it on "justified substantial doubts" (\emph{begründete erhebliche Zweifel}), a standard in every version from July 2019; what changed is how it is applied (Table F4; below).

\begin{table}[t]
\caption{Sites below the threshold in the report year, without an exception, that were entitled.}
\footnotesize
\begin{tabularx}{\linewidth}{p{0.22\linewidth}XXX}
\toprule
Entitlement year & Below threshold & Entitled & Rule in force at prognosis \\
\midrule
2021 & 201 & 0.34 & Insurers \emph{may} refute on justified substantial doubts \\
2022 & 314 & 0.64 & + COVID-19 admitted as a supporting circumstance \\
2023 & 436 & 0.71 & Refutation \emph{mandatory} on such doubts, by formal decision \\
2024 & 589 & 0.68 & + standard examples of doubt, including a missed threshold \\
\bottomrule
\end{tabularx}
\end{table}

The share of entitled below-threshold sites rose from 0.34 to 0.64 with the pandemic clause and did not fall when refutation became mandatory; with four years, a growing number of such sites (201 to 589), and the pandemic overlapping, we claim no causal effect. Whether doubts are "justified" and "substantial" is an evaluative judgement under an indeterminate legal concept; an executable rule can encode the presumption and the procedure but cannot decide the doubt.

\begin{figure}[t]
\centering
\includegraphics[width=\linewidth]{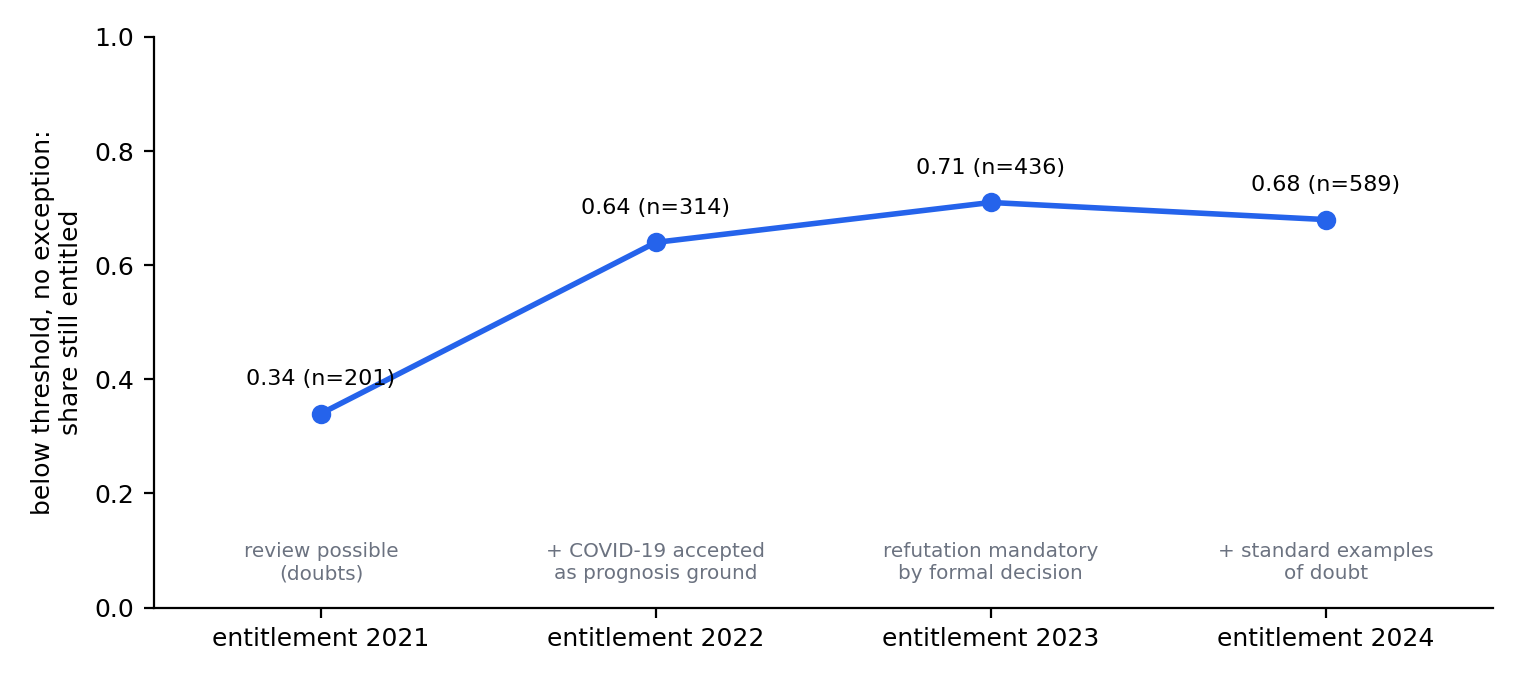}
\caption{German sites below the threshold without an exception: share still entitled, with the rule changes that preceded each entitlement year (Table F4).}
\end{figure}

\textbf{German review standard.} A further example of the Mm-R transitional clauses: very low birth-weight infants, 14 in 2021–2022, 20 in 2023, final 25. A refutation on "justified substantial doubts" is an administrative act (BSG, 25 March 2021, B 1 KR 16/20 R). Prognoses for 2022 could cite the COVID-19 pandemic (§4(2) Mm-R); from the prognoses for 2023 on, refutation is mandatory when justified substantial doubts exist and takes the form of a formal decision (§136b(5) SGB V); from 2024, standard examples of doubt, including a missed threshold, apply.

\subsection{Synthesis results}

\begin{table}[t]
\caption{Synthesis results in the comparison cases (mean over functions; n = functions, except US current law and Germany, where n counts function-years: one current-law function is scored separately in each year).}
\footnotesize
\begin{tabularx}{\linewidth}{p{0.22\linewidth}XXXXX}
\toprule
Case & Measure & No law & Current law & Statute edition only (US) & Ours (US: statute + Revenue Procedure) \\
\midrule
US & exact (±\$1), 2013–2022 & 0.970 (30) & 0.963 (30) & 0.950 (30) & 0.973 (30) \\
US & within 1\%, 2013–2022 & 1.000 (30) & 1.000 (30) & 0.986 (30) & 0.999 (30) \\
Germany & balanced accuracy, years with a changed threshold (2023 / 2024) & 0.783 / 0.825 & 0.773 / 0.810 & – & 0.773 / 0.810 \\
\bottomrule
\end{tabularx}
\end{table}

\textbf{US and Germany: the model knows the law.} In the US the pre-registered comparison (our bundle against the statute edition only) passes in one of ten years (2014), and the no-law condition is not clearly below ours (Table F5), so we make no claim for our bundle; nearly the whole gap to the statute edition comes from the 84 returns that owe the alternative minimum tax, and on the 52 returns in the personal-exemption phase-out range the synthesized rules follow the statute more closely than the independent key. In Germany the current Mm-R keeps its transitional clauses and their year-specific values, so functions built from our date-correct bundle and from the current version score identically in the years in which a threshold changed; no ordering of the German conditions survives the removal of a single function, and the case provides no evidence for retrieval (below). The sensitivity settings are those of Appendix D.2 (scripts/score\_all.py).

\textbf{Other models.} Without the amounts in memory, GPT-OSS 120B gains from retrieval in the US (within 1\% of TAXSIM on 0.574 of returns with our bundle against 0.314 with the statute edition only) and in Germany (balanced accuracy 0.855 against 0.708 for the current version). Haiku 5.5 likewise gains in the US from the Revenue Procedure (1.000 against 0.955).

\textbf{US: the model knows the amounts, and is closer to the statute than the key.} In the pre-registered comparison, statute plus Revenue Procedure (0.973) against the statute edition only (0.950), one of ten years passes the criterion (2014: 0.955 against 0.863, 9 of 9 pairs). As anticipated in the analysis plan, the no-law condition is not clearly lower than ours in any year (overall 0.970 against 0.973; 2014: 0.947, range 0.930–0.955, three functions), so we report that the model already knows the parameters and make no claim for our bundle. Functions given the statute edition only miss the alternative minimum tax on many returns; functions given our bundle compute it on every AMT return, as do two of the three current-law functions. The size of this effect can be measured. Of the 2,000 returns, 84 owe AMT. On them, functions given the statute edition only agree with the key on 0.461 on average, against 1.000 for ours, 0.972 without law, and 1.000, 1.000 and 0.274 for the three current-law functions; on the other returns every condition agrees at 0.963–0.973. Nearly the whole gap between ours and the statute edition (0.973 against 0.950) thus comes from the AMT returns. Because the store lacks §55 (Appendix F.1), the AMT amounts in the Revenue Procedure appear to cue an AMT computation the model already knows, rather than the model reading the AMT from the law; this fits the pattern that memory dominates in the US case. All remaining errors of our functions in 2013–2017 fall on the 52 returns in the personal-exemption phase-out range, where they differ from TAXSIM by \$2 to \$52: the functions apply the statute's steps of \$2,500, and TAXSIM approximates the phase-out continuously (Appendix F.1). Here the synthesized rules follow the law more closely than the independent key. This joins the AMT line and the 2023 coverage limit as a third case in which disagreement across conditions exposed a property of the key. GPT-OSS 120B, which does not have the amounts in memory, writes functions within 1\% of TAXSIM on 0.574 of returns with our bundle, 0.314 with the statute edition only, and 0.472 without law (lead over the statute edition 0.260, 95\% interval 0.246–0.274).

\textbf{Germany: the current version carries the past.} In the years in which a threshold changed, functions built from our date-correct bundle and from the current version score identically. The current Mm-R keeps its transitional clauses: it still states, for example, that a minimum volume of 10 applied to complex pancreatic surgery in 2022 and 2023 and 15 in 2024, with similar year-specific values for oesophageal surgery and very low birth-weight infants. A model reading only the current version can therefore recover the past thresholds, and version selection matters less than in Korea or the US. This pairs with the Korean observation that consolidated texts keep past addenda (Section 2). One function built with our bundle returned no value on 88\% of cases (data/de\_mm/synth/ours\_2021\_novalue.md); as fixed in the plan it is included (mean balanced accuracy over all functions 0.796; 0.863 without it). The hand-written threshold rule (0.880) chose between reported counts after looking at the key and is not a fair baseline for the LLM conditions. Functions written without any law score above both retrieval conditions (balanced accuracy 0.848 over all years, 12 function-years; 0.804 against 0.791 in the changed-threshold years), but this ordering depends on one response: counting the original no-law response that did not load as wrong on every case instead of regenerating it gives 0.774, below current law (0.811) and ours (0.796). The mean for ours is in turn pulled down by one function, the 2021 function that returned no value on 88\% of cases (balanced accuracy 0.062): by median the order is ours 0.859, no law 0.843, current law 0.835, and without that function ours averages 0.863 (data/robust\_ci.md). No ordering of the German conditions survives the removal of a single function. The model appears to know the published thresholds, so the German case provides no evidence for retrieval.

\subsection{Examples}

\emph{Following the statute where the key approximates (US).} Functions given the statute compute the personal-exemption phase-out in the statute's steps, which TAXSIM approximates continuously (Appendix F.4):

\begin{verbatim}
increments = math.ceil(excess_pe / 2500.0)
pe_reduction_pct = min(1.0, increments * 0.02)
\end{verbatim}

\emph{Memory that the text cues (US).} The US version store lacks §55, yet functions given a Revenue Procedure compute the AMT on every applicable return, and functions given only the statute edition often omit it (Appendices F.1 and F.4).

\end{appendices}

\backmatter

\bmhead{Acknowledgements}

We thank a practising lawyer, who asked to remain anonymous, for reviewing the readings behind the Korean reference rule.

\bmhead{Data and code availability}

\begin{itemize}
\item Korean statutes, decrees, annexes, and addenda are public legal texts (Korean Copyright Act Art. 7 excludes statutes from protection); we release the version store and structured annex tables. Firm-level SME labels and financial figures are derived from public DART filings; we release the derived labels with filing identifiers and the extraction scripts, so that every label can be traced to its source document.
\item US statute and Revenue Procedure texts are public domain. We release the synthetic returns and the scripts that call TAXSIM; users obtain TAXSIM35 from NBER.
\item German IQTIG result files carry no explicit licence; we release the parsing and scoring scripts and the derived thresholds, and will release the case table after confirming redistribution terms with IQTIG. Mm-R versions are linked by their G-BA identifiers.
\item All prompts, synthesized functions, agent-log audit scripts, and scoring scripts are released.
\item The release package (scripts, prompts, model responses, derived data and version stores) is available at https://github.com/nepersoned/lextemporis.
\end{itemize}

\section*{Declarations}

\bmhead{Funding} The author received no specific funding for this work.

\bmhead{Competing interests} The author declares no competing interests.

\bmhead{Ethics} The study uses public legal texts, public corporate filings, public administrative results, and synthetic taxpayers; it involves no human participants. The reviewing lawyer agreed to be acknowledged anonymously.

\bmhead{Use of AI tools} Large language models appear in two roles. As the object of evaluation, Gemini 3.8 Flash and, as further models, Claude Opus 5.5, GPT-OSS 120B, and Claude Haiku 5.5 synthesized the rules evaluated in Sections 6.3–6.4 and Appendices D and F under the isolation and audit protocol of Section 5.4. As assistants, the author used Claude (Anthropic) to propose analysis options, collect and process data, write and run analysis code, draft and translate text, check references, and review the manuscript internally, and consulted an AI-assisted legal analysis of the 2014 addendum (Section 8). The research question, the central idea of grading against independent records, the design decisions, and the author's hand-checked labels are the author's. The author is responsible for the content.

\bmhead{Author contributions} Sole author: conception, data collection, analysis, and writing.

{\small Bibliographic data for arXiv preprints and workshop papers were checked against the arXiv and conference pages, and DOIs against Crossref, on 8–9 November 2026; author lists shortened to "et al." have not been expanded.}

\end{document}